\documentclass[journal]{IEEEtran}

\ifCLASSINFOpdf
\usepackage[pdftex]{graphicx}
\graphicspath{{./}}

\DeclareGraphicsExtensions{.png,.pdf,.jpg,.jpeg}
\else
   \usepackage[dvips]{graphicx}
   \graphicspath{{Figures/}}
     \DeclareGraphicsExtensions{.png,.pdf,.jpg,.jpeg}
\fi

\usepackage{arydshln} 
\usepackage{fancyhdr}

\usepackage{algorithmic}
\usepackage{algorithm}
\usepackage{array}
\usepackage[caption=false,font=normalsize,labelfont=sf,textfont=sf]{subfig}
\usepackage{textcomp}
\usepackage{stfloats}
\usepackage{url}
\usepackage{verbatim}
\usepackage{cite}
\usepackage{multirow}
\usepackage{pifont}
\usepackage{amsmath,amssymb,amsfonts,amsbsy}
\usepackage{bm}
\usepackage{graphicx}
\usepackage{mathtools}
\usepackage{array,ragged2e}
\usepackage{booktabs}
\usepackage{makecell}

\usepackage{color}
\usepackage[table,xcdraw,dvipsnames]{xcolor}
\definecolor{myPink}{rgb}{0.9294, 0.0078, 0.5490}
\definecolor{Gray}{gray}{0.92}
\definecolor{my_color}{HTML}{E8F3F1}
\definecolor{my_color1}{HTML}{FFEACE}
\definecolor{my_color2}{HTML}{FBEAFF}
\definecolor{my_color3}{HTML}{FFC1B5}
\usepackage[
  colorlinks=true,
  linkcolor=blue,  
  citecolor=blue,  
  urlcolor=myPink,   
  linkbordercolor={1 1 1} 
]{hyperref}
\usepackage{threeparttable}
\usepackage{diagbox}
\usepackage{orcidlink}
\usepackage{pifont}

\makeatletter
\let\NAT@parse\undefined
\makeatother

\begin{document}
 
\title{Perceive-IR: Learning to Perceive Degradation Better for All-in-One Image Restoration} 

\author{
        Xu Zhang~\orcidlink{0000-0001-7685-7500},
        Jiaqi~Ma~\orcidlink{0000-0001-8491-1968},
	    Guoli~Wang~\orcidlink{0000-0001-8685-3968},          
     Qian~Zhang~\orcidlink{0009-0004-4123-8979},
     Huan~Zhang~\orcidlink{0000-0002-5507-4985},~\IEEEmembership{Member,~IEEE,} and 
        Lefei~Zhang~\orcidlink{0000-0003-0542-2280},~\IEEEmembership{Senior Member,~IEEE} 

\thanks{This work was supported by the National Natural Science Foundation of China under Grant 62431020, the Foundation for Innovative Research Groups of Hubei Province under Grant 2024AFA017, and the Fundamental Research Funds for the Central Universities under Grant 2042025kf0030.
\emph{(Xu Zhang and Jiaqi Ma contributed equally to this work.)} \emph{(Corresponding author: Lefei Zhang.)}}%

\thanks{Xu Zhang and Lefei Zhang  are with the National Engineering Research Center for Multimedia Software, School of Computer Science, Wuhan University, Wuhan 430072, China (e-mail: zhangx0802@whu.edu.cn; zhanglefei@whu.edu.cn).}

\thanks{Jiaqi Ma is with the School of Computer Science, Wuhan University, Wuhan 430072, China, and also with the Mohamed bin Zayed University of Artificial Intelligence, UAE (e-mail: jiaqima@whu.edu.cn).}

\thanks{Guoli Wang and Qian Zhang are with the Horizon Robotics, Beijing 100083, China (e-mail: guoli.wang@horizon.cc; qian01.zhang@horizon.ai).}

\thanks{Huan Zhang is with the School of Information Engineering, Guangdong University of Technology, Guangzhou 510006, China (e-mail: huanzhang2021@gdut.edu.cn).}
}

\markboth{Journal of \LaTeX\ Class Files,~Vol.~14, No.~8, August~2021}%
{Shell \MakeLowercase{\textit{\textit{et al.}}}: A Sample Article Using IEEEtran.cls for IEEE Journals}

\maketitle

\begin{abstract} %
Existing All-in-One image restoration methods often fail to perceive degradation types and severity levels simultaneously, overlooking the importance of fine-grained quality perception. 
Moreover, these methods often utilize highly customized backbones, which hinder their adaptability and integration into more advanced restoration networks.
To address these limitations, we propose Perceive-IR, a novel backbone-agnostic All-in-One image restoration framework designed for fine-grained quality control across various degradation types and severity levels. Its modular structure allows core components to function independently of specific backbones, enabling seamless integration into advanced restoration models without significant modifications.
Specifically, Perceive-IR operates in two key stages: 1) multi-level quality-driven prompt learning stage, where a fine-grained quality perceiver is meticulously trained to discern three-tier quality levels by optimizing the alignment between prompts and images within the CLIP perception space. This stage ensures a nuanced understanding of image quality, laying the groundwork for subsequent restoration; 2) restoration stage, where the quality perceiver is seamlessly integrated with a difficulty-adaptive perceptual loss, forming a quality-aware learning strategy. This strategy not only dynamically differentiates sample learning difficulty but also achieves fine-grained quality control by driving the restored image toward the ground truth while pulling it away from both low- and medium-quality samples.
Furthermore, Perceive-IR incorporates a Semantic Guidance Module (SGM) and Compact Feature Extraction (CFE). The SGM leverages semantic information from pre-trained vision models to provide high-level contextual guidance, while the CFE focuses on extracting degradation-specific features, ensuring accurate handling of diverse image degradations.
Extensive experiments demonstrate that Perceive-IR not only surpasses state-of-the-art methods but also generalizes reliably to zero-shot real-world and unknown degraded scenes, while adapting seamlessly to different backbone networks. This versatility underscores the framework's robustness and backbone-agnostic design.

\end{abstract}

\IEEEpeerreviewmaketitle
\section{Introduction} 
\label{sec:intro} 

\IEEEPARstart{I}{mage} restoration, the process of recovering a clear image from its degraded version, has seen remarkable advancements with the emergence of deep learning. Traditionally, this challenge has been addressed by task-specific networks, each specifically designed and trained to handle a unique type of degradation. This targeted approach has yielded significant success across a spectrum of restoration tasks, \textit{e.g.}, denoising \cite{DnCNN, FFDNet, ADFNet}, dehazing \cite{DehazeNet, FDGAN, DehazeFormer}, deraining \cite{UMR, MSPFN, DRSformer}, deblurring \cite{DeblurGAN, GoPro, Stripformer}, and low-light enhancement \cite{LOL, URetinex, Retinexformer}.

\begin{figure}[tbp]   %
	\centerline{\includegraphics[page=1,trim = 0mm 0mm 0mm 0mm, clip, width=1\linewidth]{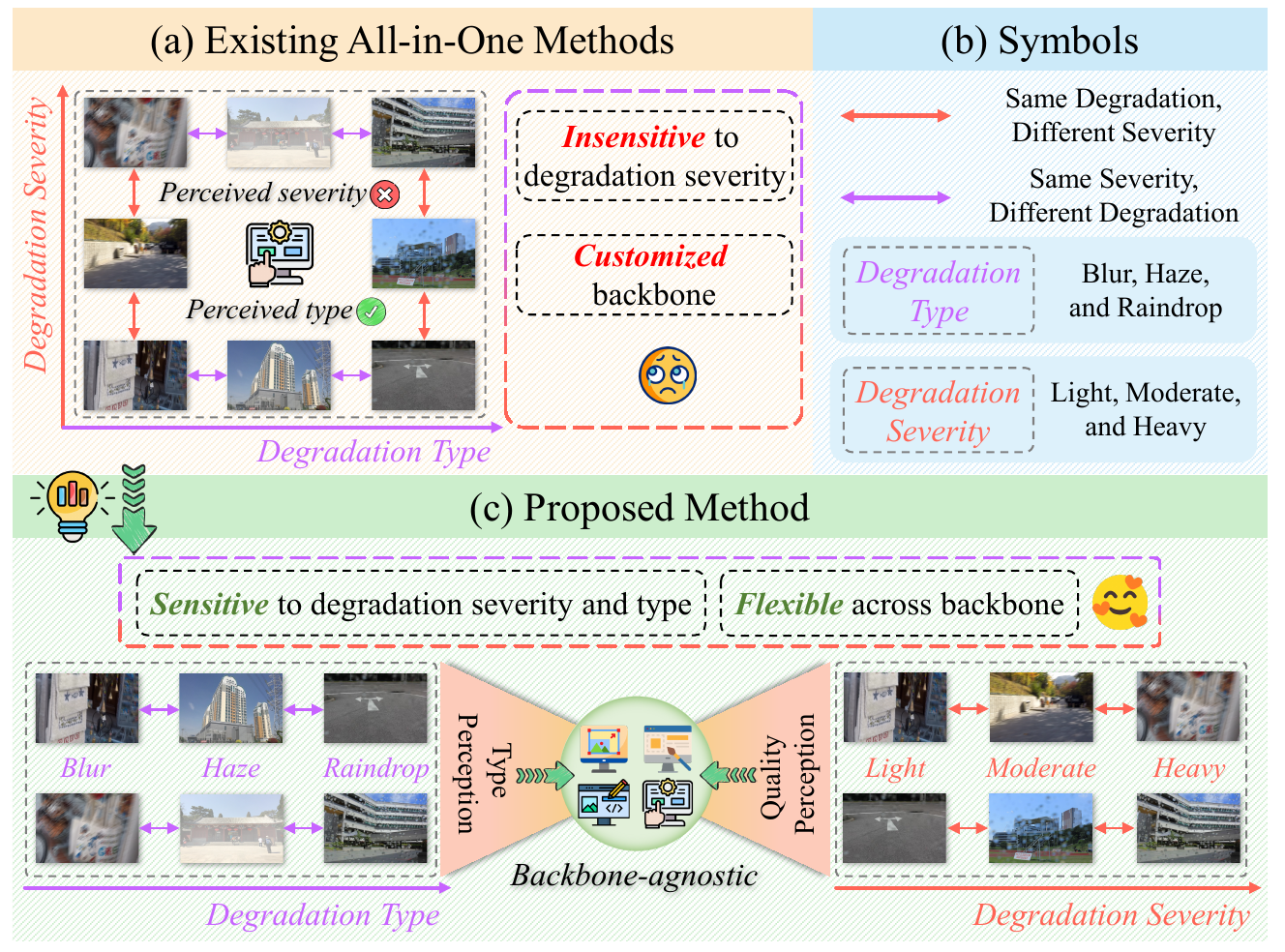}}
\vspace{-0.3em}
	\caption{{Mechanisms of existing All-in-One methods vs. our method. (a) Existing All-in-One methods are capable of recognizing degradation types (such as blur, haze, etc.) but struggle to perceive severity levels, often only distinguishing between light and heavy cases. Furthermore, their reliance on customized backbones further limits transferability. (b) Legend for symbols. (c) Our method simultaneously perceives degradation types and severity levels, while being compatible with diverse restoration backbones, offering superior flexibility and versatility.}}
	\label{fig:motivation}
	\vspace{-1em}
\end{figure}

Although task-specific methods have proven effective, their applicability is inherently restricted because they are designed to address specific types of degradation. Once switched to other degradation scenarios, the effectiveness of these methods diminishes significantly. To overcome these limitations, general image restoration methods\cite{MPRNet, MIRNet_v2, NAFNet, Restormer, FSNet, DiffIR} have been introduced. These methods are capable of handling various types of degradation, but they typically require separate models for each degradation type, making the inference stage resource-intensive and impractical.

Recently, All-in-One image restoration methods \cite{TAPE, AirNet, CLIP_AWR, IDR, ProRes, PromptIR, NDR, DINO_IR, UtilitiIR, DA-CLIP, Gridformer, InstructIR} have emerged as a potential solution to these challenges. These methods can handle multiple degradation types concurrently by employing various mechanisms. Early works, such as AirNet \cite{AirNet}, obtained discriminative representations of degraded features by explicitly constructing degradation encoders. Later methods, like ProRes \cite{ProRes} and PromptIR \cite{PromptIR}, further improved restoration performance by injecting visual prompt information. More recent studies \cite{CLIP_AWR, DINO_IR} have leveraged the feature representation capabilities of large-scale visual models like CLIP \cite{CLIP} and DINO \cite{DINO} to enhance texture reconstruction and maintain structural consistency. 

\begin{figure}[!tp] 
	\centerline{\includegraphics[page=1,trim = 0mm 0mm 0mm 0mm, clip, width=1\linewidth]{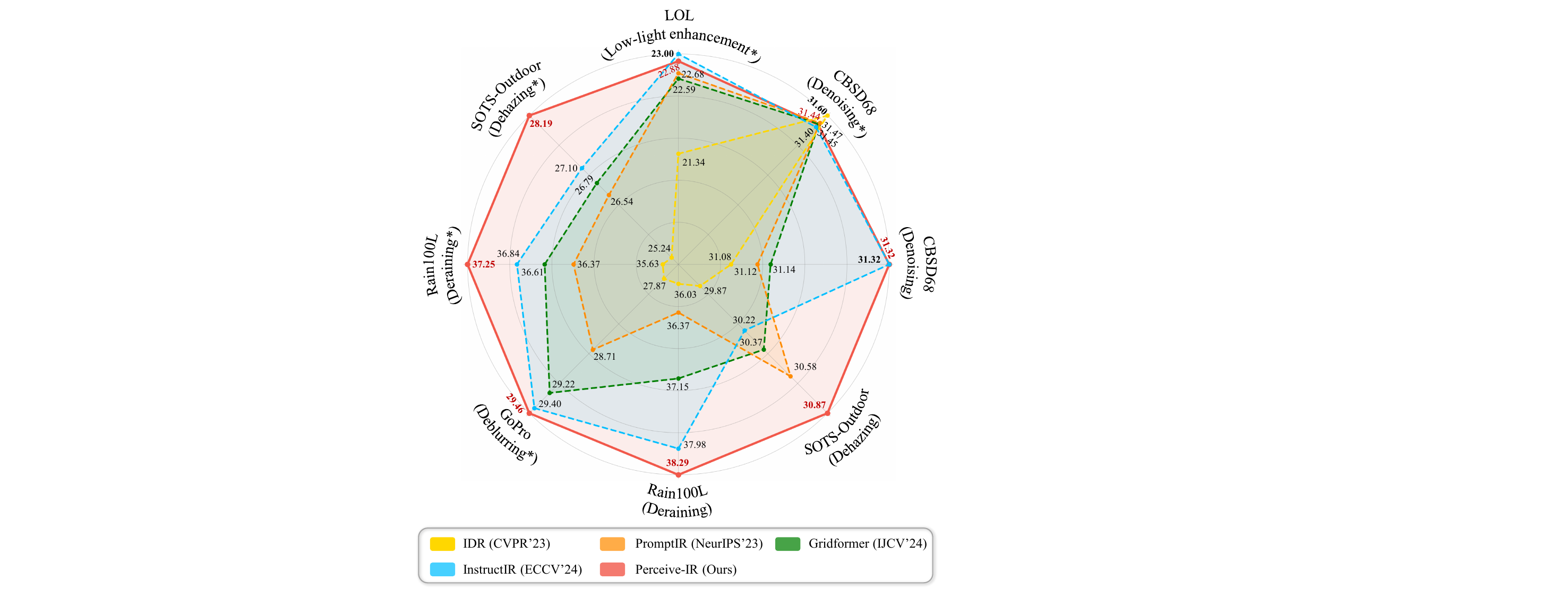}}
\vspace{-0.3em}
	\caption{PSNR comparisons with state-of-the-art All-in-One methods across two common All-in-One image restoration scenarios. * denotes results obtained under All-in-One (``Noise+Haze+Rain+Blur+Low-light'') training setting, while unmarked results are from All-in-One (``Noise+Haze+Rain'') training setting. Our method's results are marked in red, while the best results are indicated in bold.}
	\label{fig:radar}
	\vspace{-1em}
\end{figure}

Despite these advances, a significant challenge remains in accurately mapping degraded images, which exhibit varying degrees of distortion, to their corresponding ground truths. This challenge is particularly pronounced in terms of fine-grained quality perception and the ability to adapt to diverse degradation severities. As illustrated in Fig. \ref{fig:motivation}, existing All-in-One restoration methods are capable of recognizing degradation types but often struggle to perceive the severity levels of degradation. These methods typically rely on customized backbones, which limit their transferability. In contrast, our proposed method is sensitive to both degradation severity and type while being flexible across different restoration models. 

To achieve this, we introduce two key insights. First, we propose a quality-aware learning strategy that enables fine-grained quality control and dynamic adaptation to varying degradation levels. Second, we incorporate a Semantic Guidance Module (SGM) and Compact Feature Extraction (CFE). The SGM leverages semantic cues from large-scale vision models to provide high-level contextual guidance, while the CFE focuses on extracting degradation-specific features, enabling the model to perceive different types of degradation.

As demonstrated in Fig. \ref{fig:radar}, Perceive-IR outperforms existing state-of-the-art All-in-One restoration methods, exhibiting superior performance across various degradation scenarios.

Our contributions can be summarized as follows:

\noindent 
\ding{113}~
We introduce Perceive-IR, a versatile All-in-One image restoration framework that not only perceives both degradation types and fine-grained severity levels but also achieves remarkable transferability to various restoration models. %

\noindent 
\ding{113}~
We design the quality-aware learning strategy, which is based on a CLIP-aware loss and difficulty-adaptive perceptual loss, to enable fine-grained quality control and dynamic differentiation of sample learning difficulty. This strategy drives the restored image toward the ground truth while pulling it away from both low- and medium-quality samples. %

\noindent 
\ding{113}~
We develop a novel semantic guidance module that combines a pre-trained DINO-v2 model with a prompt guidance module, generating feature representations enriched with both high-level semantic priors and degradation-specific cues, thereby significantly enhancing the restoration process.

\noindent 
\ding{113}~Extensive results demonstrate that Perceive-IR achieves state-of-the-art performance across diverse image restoration tasks. Importantly, it is inherently backbone-agnostic, seamlessly adapting to various restoration models, making it a highly flexible framework for All-in-One image restoration. %

\section{Related Work} 

\subsection{All-in-One Restoration} 
All-in-One restoration \cite{TAPE, AirNet, CLIP_AWR, IDR, ProRes, PromptIR, NDR, DINO_IR, UtilitiIR, DA-CLIP, Gridformer, InstructIR}, which aims to recover clean images from multiple degradations through a unified model, has grown to be a promising field of low-level vision tasks. Compared to task-specific \cite{DnCNN, FFDNet, AFDformer, XiaoYi, LURE, DehazeNet, FDGAN, DehazeFormer, UMR, MSPFN, DRSformer, DeblurGAN, GoPro, Stripformer, LOL, Retinexformer, URetinex} and general \cite{MPRNet, MIRNet_v2, NAFNet, DGUNet, Restormer, FSNet, DiffIR, MambaIR} image restoration, All-in-One restoration is more advantageous in terms of model storage efficiency and practical applications. The main challenge lies in using a single set of model parameters to handle various types of degradation and accurately restore the corresponding components. To achieve this, AirNet \cite{AirNet} proposed learning discriminative degradation representation using contrastive learning. IDR \cite{IDR} took a different approach by  utilizing a two-stage ingredients-oriented restoration network. PromptIR \cite{PromptIR} and ProRes \cite{ProRes} further enhanced the network's ability to handle multiple degradation through vision prompts. More recently, CLIP-AWR \cite{CLIP_AWR}, DA-CLIP \cite{DA-CLIP}, and DINO-IR \cite{DINO_IR} leveraged pre-trained large-scale vision models to excel in All-in-One restoration tasks. 

However, in practice, degraded images often suffer from varying levels of corruption. The above methods may fall into the trap of processing these images with the same restoration effort, resulting in sub-optimal performance. Based on this observation, Chen \textit{et al.} \cite{UtilitiIR} designed a blind All-in-One image restoration method by learning an image quality ranker. However, it depends on the simple image quality metric, \textit{i.e.}, PSNR, and is susceptible to variations in certain parameters. 
Furthermore, most of these methods rely on highly customized backbones, which compromises model flexibility. In contrast, our approach incorporates a quality-aware learning strategy and compact feature extraction to achieve fine-grained quality control and uncover degradation-specific cues. This demonstrates superior sensitivity to both the severity and type of degradation, while maintaining remarkable adaptability across various restoration models.

\subsection{Prompt Learning} 
In recent years, prompt learning as an emerging learning paradigm has seen significant advancements in the field of natural language processing \cite{brown2020language}. Its effectiveness has led to its widespread application in vision-related tasks \cite{jia2022visual}. The core idea is to enable pre-trained models to better understand and perform downstream tasks by constructing specific prompts. In natural language processing, these prompts are usually given in the form of text, whereas in computer vision, they may involve images, text, or other forms of data.
Recently, CLIP-LIT \cite{CLIP_LIT} has shown that initializing textual prompts in CLIP can aid in extracting more accurate low-level image representations. 
However, relying solely on distinguishing between positive and negative prompts may fail to effectively guide the restoration process towards achieving undistorted images. This is because the vast spectrum of possibilities between the two prompts can complicate the accurate mapping of degraded images to their pristine counterparts.
To address this, we propose a multi-level quality-driven prompt learning stage that further refines the differentiation among three-tier prompt pairs. This approach enables more nuanced and precise control over the restoration process, ensuring a more accurate alignment with undistorted images.

\begin{figure*}[!htp]  
	\centerline{\includegraphics[page=1,trim = 0mm 0mm 0mm 0mm, clip, width=0.9\linewidth]{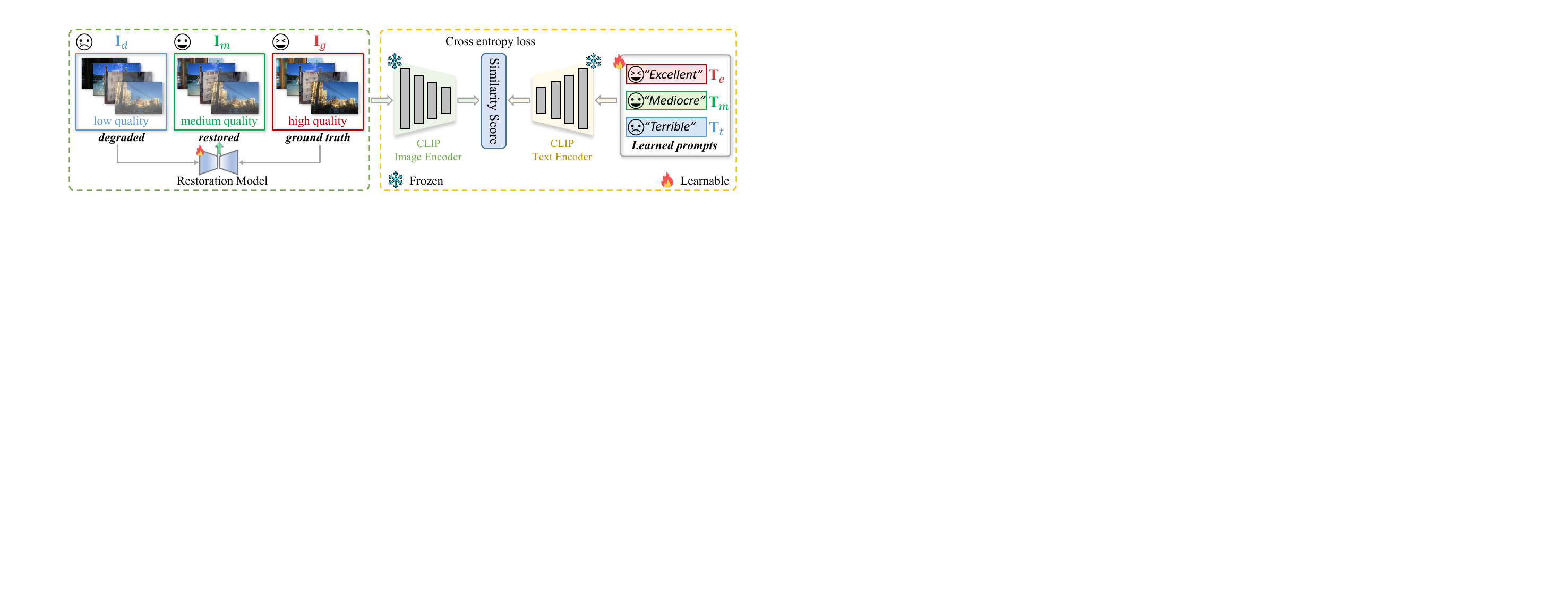}}
\vspace{-0.3em} 
	\caption{In the proposed multi-level quality-driven prompt learning stage, we initialize and train textual prompts using image-text pairs categorized into three tiers of quality. The medium quality images are obtained by training the restoration model (e.g., Restormer \cite{Restormer}) using a cross-validation strategy.
    Then, these image-text pairs are trained with cross-entropy loss in the CLIP model. Once trained, the learned prompts are fixed and used to guide the restoration of high-quality images during the subsequent restoration stage.}
	\label{fig:stage1}
	\vspace{-1em}
\end{figure*}

\vspace{-0.5em} 
\subsection{Large-scale Vision Models} 
Large-scale vision models (LVMs) have demonstrated powerful robust feature representation and zero-shot transfer capabilities across various tasks. For example, CLIP \cite{CLIP} has been widely adopted in numerous downstream applications due to its remarkable ability to align semantic information between vision and language. Similarly, self-supervised ViT models like DINO \cite{DINO} and DINO-v2 \cite{DINOv2} have proven effective in multiple domains, eliminating the need for labeled input data. Given that these LVMs can extract valuable prior knowledge from external hyper-scale datasets, leveraging pre-trained LVMs has become increasingly popular in the low-level vision community \cite{CLIP_AWR, DA-CLIP, DINO_IR}.
Inspired by these advancements, our work utilizes the semantic prior knowledge and structural information extracted by the proposed DINO-v2-based semantic guidance module to guide the restoration process.

\section{Method} 
\label{Method}
\subsection{Overview Pipeline} 
Our Perceive-IR contains two stages: a prompt learning stage and a restoration stage. In the prompt learning stage, 
we initialize and train textual prompts using image-text pairs categorized into three-tier quality by constraining the text-image similarity in the CLIP perception space. The learned prompts are fixed and then used to guide the restoration process during the subsequent stage. In the restoration stage, we use compact feature extraction (CFE) to learn distinct degradation representations. These representations are then concatenated with semantic priors extracted by pre-trained DINO-v2 encoder. These concatenated features modulate the output features of the decoder within the prompt guidance module (PGM). The calibrated features are further processed by the prior guidance cross attention (PGCA) to emphasize and extract structural semantic information and convey it. Moreover, we utilize the quality-aware learning strategy which contains CLIP-aware loss and difficulty-adaptive perceptual loss to realize fine-grained quality control and guide the restoration of high-quality images.

\subsection{Multi-level Quality-Driven Prompt Learning Stage} 

The process of multi-level quality-driven prompt learning stage is shown in Fig. \ref{fig:stage1}. We use the CLIP model to learn three types of prompts: ``terrible'', ``mediocre'', and ``excellent'', which correspond to the three-tier quality levels (low, medium, and high) of images processed by the CLIP image encoder. Specifically, we divide the degraded inputs equally into two subsets. Then, we train a restoration model, \textit{i.e.}, Restormer \cite{Restormer} to obtain the restored images $\mathbf{I}_m$ via cross-validation strategy. The restored images are regarded as medium quality images. After that, given a low quality degraded image $\mathbf{I}_d \in \mathbb{R}^{H \times W \times 3}$, a medium quality restored image $\mathbf{I}_m \in \mathbb{R}^{H \times W \times 3}$, and a high quality ground truth $\mathbf{I}_g \in \mathbb{R}^{H \times W \times 3}$, we randomly initialize a ``terrible'' textual prompt $\mathbf{T}_t \in \mathbb{R}^{N \times 512}$, a ``mediocre'' textual prompt $\mathbf{T}_m \in \mathbb{R}^{N \times 512}$, and an ``excellent'' textual prompt $\mathbf{T}_e \in \mathbb{R}^{N \times 512}$. $N$ represents the number of embedded tokens in each prompt. Then, we feed the low, medium, and high quality images $\mathbf{I}_d$, $\mathbf{I}_m$, and $\mathbf{I}_g$ to the CLIP image encoder to obtain their latent encoding. Meanwhile, we extract the latent encodings of the ``excellent'', ``mediocre'', and ``terrible'' textual prompts by feeding them to the CLIP text encoder. Based on the text-image similarity in the CLIP latent space, we use the cross entropy loss $\mathcal {L}_{ce}$  of classifying the low, medium, and high quality images to learn these prompts. The $\mathcal {L}_{ce}$ can be described as
\begin{equation}
\begin{split}  
\mathcal {L}_{ce} =& -\frac{1}{3}\sum_{i\in\{d, m, g\}}\sum_{j\in\{e, m, t\}}\mathcal{Y}_{ij}\log  \\
 &\Bigg(\frac{\exp\big(\mathcal{S}\big(\Phi_{\mathcal I}(\mathbf{I}_{i}), \Phi_{\mathcal T}(\mathbf{T}_{j})\big)\big)}{\sum_{k\in\{e, m, t\}}
 \exp\big(\mathcal{S}\big(\Phi_{\mathcal I}(\mathbf{I}_{i}), \Phi_{\mathcal T}(\mathbf{T}_{k})\big)\big)}\Bigg), 
\end{split}
\end{equation}
where $\mathcal{S}(\cdot,\cdot)$ denotes cosine similarity and $\mathcal{Y}_{ij}$ is the label of the current image $\mathbf{I}_{i}$; $\Phi_{\mathcal I}(\cdot)$ and $\Phi_{\mathcal T}(\cdot)$ denote CLIP image encoder and CLIP text encoder, respectively.

\begin{figure*}[!htp]  
	\centerline{\includegraphics[page=1,trim = 0mm 0mm 0mm 0mm, clip, width=1\linewidth]{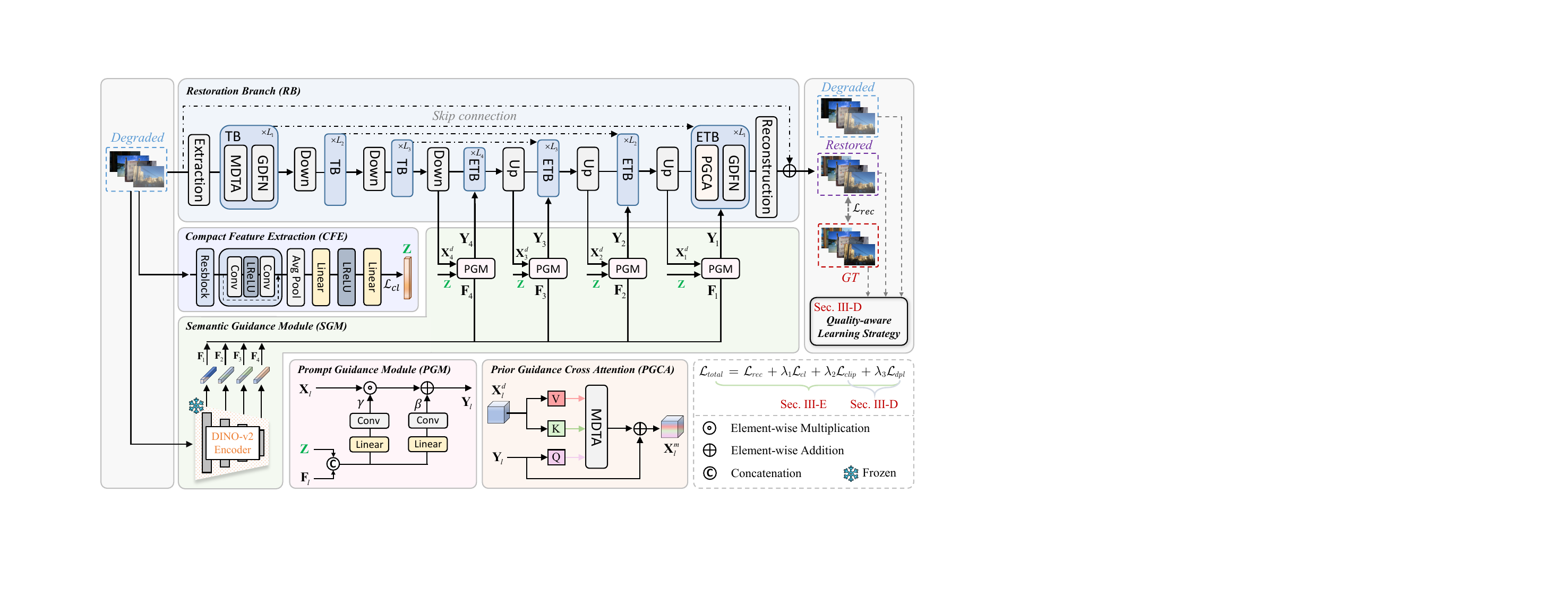}}
\vspace{-0.3em} 
	\caption{The proposed restoration stage consists of: (a) Restoration Branch (RB): A 4-level U-shaped encoder-decoder structure that incorporates Transformer Block (TB) \cite{Restormer} in the encoder and Enhanced Transformer Block (ETB) in the decoder. (b) Compact Feature Extraction (CFE): A module designed to generate distinctive degradation representation. (c) Semantic Guidance Module (SGM): Comprising a pre-trained DINO-v2 \cite{DINOv2} and the Prompt Guidance Module (PGM) to produce feature representations enriched with semantic and degradation priors.}  
	\label{fig:stage2}
	\vspace{-1em}
\end{figure*}

\subsection{Restoration Stage} 

\subsubsection{Restoration Branch} 
As shown in Fig. \ref{fig:stage2}, given a degraded input $\mathbf{I}_d \in \mathbb{R}^{H \times W \times 3}$, we first apply a 3$\times$3 convolution
to extract shallow embeddings $\mathbf{X}_s \in \mathbb{R}^{H \times W \times C}$, where H$\times$W denotes the spatial dimension and $C$ is the number of channels. Next, the shallow features $\mathbf{X}_s$  are gradually and hierarchically encoded into deep features $\mathbf{X}_l^{e,d} \in \mathbb{R}^{\frac {H}{2^{l-1}} \times \frac {W}{2^{l-1}} \times 2^{l-1}C}$. After encoding the degraded input into latent features $\mathbf{X}_4 \in \mathbb{R}^{\frac {H}{8} \times \frac {W}{8} \times 8C}$, the decoder progressively recovers the high-resolution representations. Finally, a 3 × 3 convolution is applied to reconstruct the image $\mathbf{I}_r \in \mathbb{R}^{H \times W \times 3}$. 

In the restoration branch, we choose Restormer \cite{Restormer} as backbone. Specifically, in the encoder layer, we integrate Multi-Dconv Head Transposed Attention (MDTA) \cite{Restormer} and Gated-Dconv Feed-Forward Network (GDFN) \cite{Restormer} to jointly construct the Transformer Block (TB). In the latent and decoder layers, we introduce the Prior Guidance Cross Attention (PGCA) and GDFN to jointly construct the Enhanced Transformer Block (ETB). The process of restoration branch can be described as 
\begin{equation}
\mathcal{F}_{tb}(\mathbf{X}_l^e)  =  \mathcal{F}_{gdfn}\big(\mathcal{F}_{mdta}(\mathbf{X}_l^e)\big),
\end{equation}
\begin{equation}
\mathcal{F}_{etb}(\mathbf{X}_l^d, \mathbf{Y}_l)  =  \mathcal{F}_{gdfn}\big(\mathcal{F}_{pgca}(\mathbf{X}_l^d, \mathbf{Y}_l)\big),
\end{equation}
\begin{equation}
\mathbf{I}_{r} =Conv\bigg(\mathcal{F}_{etb}^{4,3,2,1}\Big(\mathcal{F}_{tb}^{1,2,3}\big(Conv(\mathbf{I}_d)\big){\downarrow}_{\times2},\mathbf{Y}_l\Big){\uparrow_{\times2}}\bigg) + \mathbf{I}_d,
\end{equation}
where $\mathcal{F}_{gdfn}(\cdot)$, $\mathcal{F}_{mdta}(\cdot)$, $\mathcal{F}_{pgca}(\cdot)$, $\mathcal{F}_{tb}(\cdot)$, and $\mathcal{F}_{etb}(\cdot)$ indicate GDFN, MDTA, PGCA, TB, and ETB processes, respectively; $\mathbf{X}_l^e$ denotes the output of $l$-th encoder layer and $\mathbf{X}_l^d$ represents the input of $l$-th decoder layer, respectively; $Conv(\cdot)$ indicates 3$\times$3 convolution operation and $\mathbf{Y}_l$ denotes output of prompt guidance module; $\downarrow$ and $\uparrow$ indicate down-sampling and up-sampling, respectively.

\subsubsection{Compact Feature Extraction} In the restoration stage, we propose Compact Feature Extraction (CFE) to extract a compact degradation representation $\mathbf{Z} \in \mathbb{R}^{1 \times 128}$. Then, we utilize the degradation loss $\mathcal{L}_{cl}$ (as detailed in Sec. \ref{sec:total_loss}) to optimize CFE by leveraging the consistency of images with the same degradation and the inconsistency across different degradation. After that, $\mathbf{Z}$ will be concatenated with the multi-scale features $\mathbf{F}_l$ extracted by pre-trained large vision model. The deep feature $\mathbf{X}_l^d$ and the concatenated feature will perform affine transformation in Prompt Guidance Module (PGM) to obtain $\mathbf{Y}_l$. Subsequently, the $\mathbf{Y}_l$ is employed to execute cross-attention within the Enhanced Transformer Block (ETB).

\subsubsection{Semantic Guidance Module} 
Recently, large-scale vision models (\textit{e.g.}, DINO family \cite{DINO, DINOv2} ) have demonstrated their potential in a series of visual downstream tasks in a self-supervised manner. DINO-v2, as an improved version of DINO, can provide more powerful feature representation thanks to pre-training on more than one million data. To this end, we employ the DINO-v2 to extract useful semantic feature priors from degraded images. Additionally, a prompt guidance module (PGM) is designed to help the network better capture and preserve structural semantic information. As shown in Fig. \ref{fig:stage2}, given a degraded image $\mathbf{I}_d$, it goes through the pre-trained DINO-v2 encoder and outputs four different levels of semantic features $\mathbf{F}_l \in \mathbb{R}^{1 \times 768}$ ($l$ = 1, 2, 3, 4). This process can be described as 
\begin{equation}
\mathbf{F}_1, \mathbf{F}_2, \mathbf{F}_3, \mathbf{F}_{4} = \mathcal{D}_{i}\bm(\mathbf{I}_d\bm),\ i = 1, 4, 8, 12,
\end{equation}
where $\mathcal{D}_{i}(\cdot)$ denotes the $i$-th layer of DINO-v2 encoder. Then, the semantic prior and learned degradation representation $\mathbf{Z}$ are exploited to generate reliable content to guide the restoration by PGM. The modulated feature $\mathbf{Y}_l$ is transmitted to prior guidance cross attention (PGCA) to guide the restoration branch. Specifically, we take $\mathbf{Y}_l$ as query to perform the cross-attention. The above process can be expressed as 
\begin{equation}
\mathbf{Y}_{l}  =  \mathcal{F}_{pgm}(\mathbf{F}_{l}, \mathbf{Z}, \mathbf{X}_{l}^{d}),
\end{equation}
\begin{equation}
CrossAtt(\mathbf{Y}_l,\mathbf{X}_l^d, \mathbf{X}_l^d) = {Softmax}(\mathbf{\hat{Q}} \otimes \mathbf{\hat{K}}/\alpha \big) \otimes \mathbf{\hat{V}},
\end{equation}
\begin{equation}
\mathbf{X}_l^m  =  \mathcal{F}_{mdta}\big(CrossAtt\bm(\mathbf{Y}_l, \mathbf{X}_l^d, \mathbf{X}_l^d\bm)\big) + \mathbf{Y}_l,
\end{equation}
where $\mathcal{F}_{pgm}$ indicates PGM process; $\otimes$ denotes matrix multiplication.

\subsection{Quality-aware Learning Strategy} 
\subsubsection{CLIP-aware Loss} 
\begin{figure}[!tp]  
	\centerline{\includegraphics[page=1,trim = 0mm 0mm 0mm 0mm, clip, width=1\linewidth]{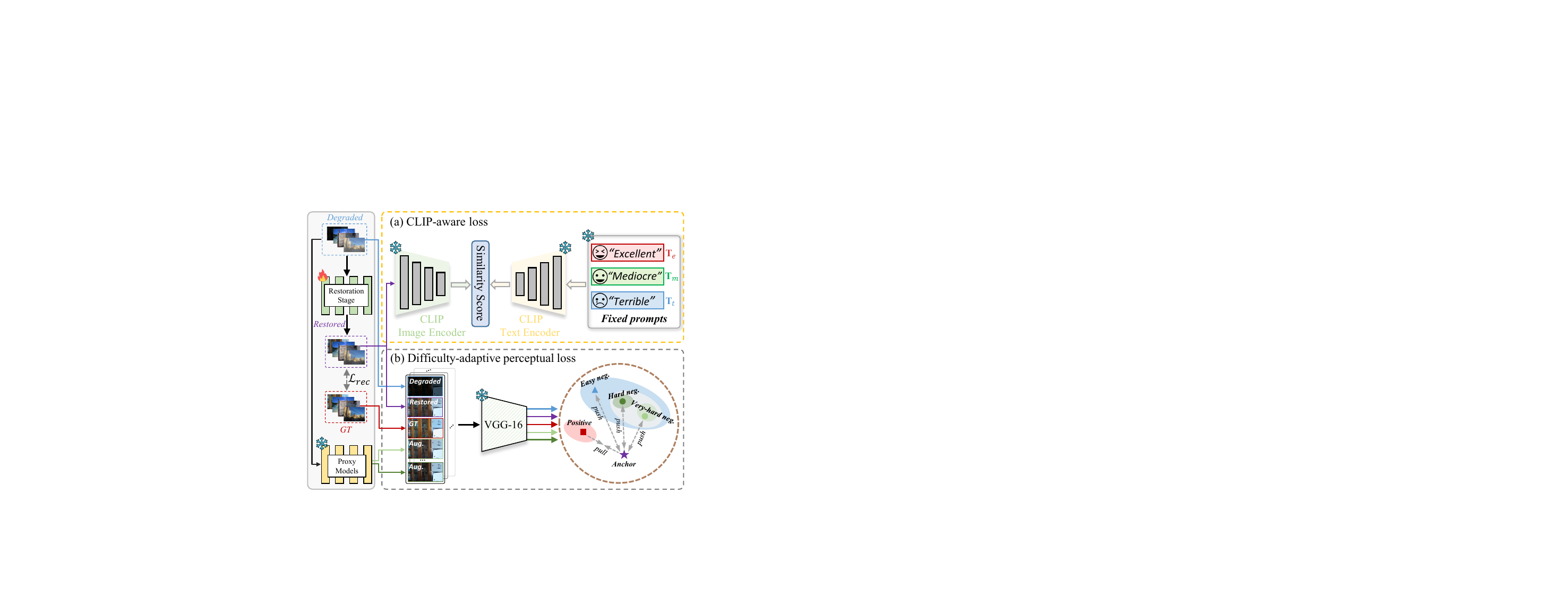}}
	\caption{The proposed quality-aware learning strategy contains two components: (a) The CLIP-aware loss penalizes the dissimilarity between the restored image and the ``excellent" prompt, guiding the restored image to better resemble the ground truth; (b) The difficulty-adaptive perceptual loss dynamically adjusts its behavior based on the difficulty level of the restoration process by distinguishing the learning difficulty of samples in feature space.}  
	\label{fig:loss}
	\vspace{-1em}
\end{figure}

As illustrated in Fig. \ref{fig:loss}, given the learned text prompts obtained from the prompt learning stage, we first fix these prompts and then train the restoration model using the CLIP-aware loss $\mathcal{L}_{clip}$. The $\mathcal{L}_{clip}$ can be described as
\begin{equation}
    \label{eq2}\mathcal{L}_{clip}=1-\frac{\exp\Big(\mathcal{S}\big(\Phi_{\mathcal I}(\mathbf{I}_{r}), \Phi_{\mathcal T}(\mathbf{T}_{e})\big)\Big)}{\sum_{k\in\{e, m, t\}}\exp\Big(\mathcal{S}\big(\Phi_{\mathcal I}(\mathbf{I}_{r}),\Phi_{\mathcal T}(\mathbf{T}_{k})\big)\Big)}, 
\end{equation}
where $\mathcal{S}(\cdot,\cdot)$ denotes cosine similarity. The $\mathcal{L}_{clip}$ is to minimize the gap between the restored images $\mathbf{I}_{r}$ and the ``excellent'' prompt $\mathbf{T}_{e}$, while maximizing the gap between $\mathbf{I}_{r}$ with the ``mediocre'' prompt $\mathbf{T}_{m}$ and ``terrible'' prompt $\mathbf{T}_{t}$, respectively.
It emphasizes pulling the restored results closer to the ground truth, guiding the model to generate high-quality images.

\subsubsection{Difficulty-adaptive Perceptual Loss} 
Using contrastive regularization has been demonstrated as an effective way of improving image restoration performance \cite{CR, C2P}. Based on this, we introduce the difficulty-adaptive perceptual loss to dynamically adjust its behavior based on the difficulty level of the restoration process. It effectively handles both easy and hard samples, ensuring that the network focuses appropriately on challenging samples while efficiently processing easy cases. As shown in Fig. \ref{fig:loss}, the image predicted $\mathbf{I}_{r}$ by the network serves as the anchor, the reference image $\mathbf{I}_{g}$ is used as positive sample, the degraded $\mathbf{I}_{d}$ is used as easy negative sample, and the augmented images $\mathbf{I}_{q}$ from proxy restoration models (\textit{e.g.}, InstructIR \cite{InstructIR}, PromptIR \cite{PromptIR}, FSNet \cite{FSNet}, and MambaIR \cite{MambaIR}) as hard or very-hard negative samples.  

Specifically, at the beginning of the $k$-th epoch, we evaluate the performance of network by calculating the average PSNR score and categorize the non-easy negative sample as a very-hard sample if its PSNR surpasses the current network performance, otherwise, it is classified as hard negative sample. Then, the weight $\mathcal{O}_{k}$ of the $k$-th epoch assigned to a non-easy negative sample $\mathcal{N}_{q}$, is defined as 
\begin{equation}
\mathcal{O}_{k}(\mathcal{N}_{q})=\left\{\begin{array}{l}
1 +\gamma, \quad \text{avgPSNR}(\mathbf{I}_{r}, \Theta_{k-1}) \geq \text{PSNR}(\mathcal{N}_{q}), \\
1-\gamma, \quad \text{otherwise},
\end{array}\right.
\end{equation}
where $q$ = 1, 2, $\cdots$, $z$; $z$ is the number of non-easy negatives, and $\gamma$ is a hyperparameter set to 0.25; $\Theta_{k-1}$ represents the model weights from the ($k-1$)-th epoch. The weights of hard and very-hard negatives are set to $1 + \gamma$ and $1-\gamma$, respectively. Based on this, the proposed $\mathcal {L}_{dpl}$ can be written as 
\begin{equation}
\mathcal {L}_{dpl} =\sum_{i=\text{3,7,11,15}} \xi _{i}\frac{\mathcal{M}_{i}(\mathbf{I}_{g}, \mathbf{I}_{r})}{\lambda \mathcal{M}_{i}(\mathbf{I}_{d}, \mathbf{I}_{r})+{\textstyle \sum_{q=1}^{z}} \mathcal{O}_{k}(\mathcal{N}_{q}) \mathcal{M}_{i}(\mathcal{N}_{q}, \mathbf{I}_{r})},
\end{equation}
where ${\mathcal{M}_i(\cdot, \cdot)}=\left \|{\mathcal{V}_{i}(\cdot)-\mathcal{V}_{i}(\cdot)}  \right \|_{1}$; $\mathcal{V}_i(\cdot)$ represents the $i$-th latent feature extracted from the pre-trained VGG-16. We use the latent features of 3rd, 7-th, 11-th, 15-th layers from VGG-16 to calculate ${\mathcal{M}_i(\cdot, \cdot)}$. We assign a fixed weight $\lambda$ to the easy negative and set $\lambda$ = 2. $\left \{ \xi_i \right \}$ represents the set of hyperparameters. We set $\xi_i$ ($i$ = 3, 7, 11, 15) to $\frac{1}{12}$, $\frac{1}{6}$, $\frac{1}{3}$, and 1, respectively.

\subsection{Training Loss}  
\label{sec:total_loss}
In the prompt learning stage, we use $\mathcal{L}_{ce}$ to learn the initial text prompt pairs.
In the restoration stage, we first utilize Compact Feature Extraction (CFE) and introduce a degradation-aware loss $\mathcal{L}_{cl}$ to learn degradation representations. 
The $\mathcal {L}_{cl}$ can be written as
\begin{equation}
\mathcal {L}_{cl} = - \log \frac{\exp\left({\mathcal{S}(\mathbf{Z}, \mathbf{Z}^{+}})/\tau\right)}{\exp\left({\mathcal{S}(\mathbf{Z}, \mathbf{Z}^{+}})/\tau\right)+\sum_{i=1}^{n} \exp\left({\mathcal{S}(\mathbf{Z}, \mathbf{Z}^{-}_{i})}/\tau\right)} ,
\end{equation}
where $n$ denotes the number of negative samples, and $\tau$ indicates the temperature, which is set to 0.1. 
Given several images containing different degradation types, we randomly crop two patches of the same resolution on an image and let them be positive samples, while letting patches of different degradation types be negative samples. Then, feed them into CFE to get the anchor feature $\mathbf{Z}$, the positive feature $\mathbf{Z}^{+}$, and the negative feature $\mathbf{Z}^{-}$. 

The total loss of the restoration stage can be described as
\begin{equation}
\label{Eq_totalloss}
\mathcal {L}_{total} = \mathcal{L}_{rec} + \lambda_{1}\mathcal{L}_{cl} + \lambda_{2}\mathcal{L}_{clip} + \lambda_{3}\mathcal {L}_{dpl}, 
\end{equation}
where $\mathcal {L}_{rec}$ is the $\mathcal {L}_{1}$ norm loss; $\lambda_{1}$, $\lambda_{2}$, and $\lambda_{3}$ are hyperparameters to make a balance of the total loss.

\begin{table*}[tbp]
\caption{Dataset summary under two training settings.}
\label{tab:dataset}
\centering
  \renewcommand\arraystretch{1.15}
    \resizebox{\linewidth}{!}{\begin{tabular}{c|c|c|c}
    \toprule[1pt]
\toprule[0.5pt]

\bf Setting & {\bf Degrdation} & {\bf Training dataset (Number)} & \bf Testing dataset (Number) \\ 

\hline

\multirow{5}{*}{\rotatebox{90}{\bf{One-by-One}}} \multirow{5}{*}{\rotatebox{90}{{(Single-task)}}} 
& {Noise (\textcolor{red}{\textbf{N}})} 
& {\textcolor{red}{\textbf{N}}: BSD400 \cite{BSD400}+WED \cite{WED} (400+4744)} 
& \textcolor{red}{\textbf{N}}: CBSD68 \cite{BSD68}+Urban100 \cite{Urban100}+Kodak24 \cite{Kodak24} (68+100+24)
\\  

& {Haze (\textcolor{blue}{\textbf{H}})} 
& \textcolor{blue}{\textbf{H}}: RESIDE-$\beta$-OTS \cite{SOTS} (72135) 
& \textcolor{blue}{\textbf{H}}: SOTS-Outdoor \cite{SOTS} (500)  
\\

& Rain (\textcolor{green}{\textbf{R}}) 
& \textcolor{green}{\textbf{R}}: Rain100L \cite{Rain100L} (200) 
& \textcolor{green}{\textbf{R}}: Rain100L \cite{Rain100L} (100) 
\\

& Blur (\textcolor{purple}{\textbf{B}})
& \textcolor{purple}{\textbf{B}}: GoPro \cite{GoPro} (2103)
& \textcolor{purple}{\textbf{B}}: GoPro \cite{GoPro} (1111)  
\\  

& Low-Light (\textcolor{pink}{\textbf{L}})
& \textcolor{pink}{\textbf{L}}: LOL \cite{LOL} (485)
& \textcolor{pink}{\textbf{L}}: LOL \cite{LOL} (15)  
\\

\hline
\hline

\multirow{8}{*}{\rotatebox{90}{\textbf{All-in-One}}} \multirow{8}{*}{\rotatebox{90}{{(Multi-task)}}} 
& \multirow{3}{*}{\textcolor{red}{\textbf{N}}+\textcolor{blue}{\textbf{H}}+\textcolor{green}{\textbf{R}}} 
& \multirow{3}{*}{\makecell{BSD400\cite{BSD400}+WED\cite{WED}+RESIDE-$\beta$-OTS\cite{SOTS}+Rain100L\cite{Rain100L}\\ \textbf{Number}: 400+4744+72135+200\\ \textbf{Total}: 77479}} 
& \textcolor{red}{\textbf{N}}: CBSD68 \cite{BSD68} (68) 
\\ 

&  
&  
& \textcolor{blue}{\textbf{H}}: SOTS-Outdoor \cite{SOTS} (500) 
\\  
 
&  
&  
& \textcolor{green}{\textbf{R}}: Rain100L \cite{Rain100L} (100) 
\\ 

\cdashline{2-4}

& \multirow{5}{*}{\bf {\textcolor{red}{\textbf{N}}+\textcolor{blue}{\textbf{H}}+\textcolor{green}{\textbf{R}}+\textcolor{purple}{\textbf{B}}+\textcolor{pink}{\textbf{L}}}} 
& \multirow{5}{*}{\makecell{BSD400\cite{BSD400}+WED\cite{WED}+RESIDE-$\beta$-OTS\cite{SOTS}+Rain100L\cite{Rain100L}\\+GoPro\cite{GoPro}+LOL\cite{LOL}\\ \textbf{Number}: 400+4744+72135+200+2103+485\\ \textbf{Total}: 80067}} 
& \textcolor{red}{\textbf{N}}: CBSD68 \cite{BSD68} (68) 
\\ 

& 
& 
& \textcolor{blue}{\textbf{H}}: SOTS-Outdoor \cite{SOTS} (500)
\\  

&  
& 
&  \textcolor{green}{\textbf{R}}: Rain100L \cite{Rain100L} (100)
\\  

&  
& 
&  \textcolor{purple}{\textbf{B}}: GoPro \cite{GoPro} (1111)
\\ 

&  
& 
&  \textcolor{pink}{\textbf{L}}: LOL \cite{LOL} (15) 
\\ 

    \bottomrule[1pt]
    \end{tabular}}
    \vspace{-1em}
\end{table*}

\section{Experiments}
\subsection{Experimental Setup} 
\subsubsection{Datasets} Following \cite{AirNet, PromptIR, IDR}, we differentiate the training setups into ``All-in-One'' and ``One-by-One'' based on whether datasets are combined for mixed training. In our study, we summarise two common settings of mixed degradations:  {``Noise+Haze+Rain (\textcolor{red}{\bf{N}}+\textcolor{blue}{\bf{H}}+\textcolor{green}{\bf{R}})''} and {``Noise+Haze+Rain+Blur+Low-light (\textcolor{red}{\textbf{N}}+\textcolor{blue}{\textbf{H}}+\textcolor{green}{\bf{R}}+\textcolor{purple}{\textbf{B}}+\textcolor{pink}{\textbf{L}})''}. Under the ``One-by-One'' setting, the model is trained and tested using datasets from a single image restoration task at a time. Under the ``All-in-One'' setting, the model is trained and tested using mixed datasets from multiple image restoration tasks. 
In Tab. \ref{tab:dataset}, we detail the training and testing datasets used in our experiments. For single-task image restoration, \textbf{i)} Image denoising: we conduct training using a merged dataset of BSD400 \cite{BSD400} and WED \cite{WED} with 400 and 4,744 clear images, respectively. Noisy images are generated with Gaussian noise ($ \sigma \in \{15, 25, 50\} $). Testing is performed on CBSD68 \cite{BSD68}, Urban100 \cite{Urban100}, and Kodak24 \cite{Kodak24} datasets. \textbf{ii)} Image dehazing: we use the OTS dataset of RESIDE-$\beta$ \cite{SOTS} with 72,135 pairs for training and 500 images from SOTS-Outdoor \cite{SOTS} dataset for testing. \textbf{iii)} Image deraining: we use the Rain100L \cite{Rain100L} dataset with 200 pairs of images for training and 100 pairs for testing. \textbf{iv)} Image deblurring: we train the model on the GoPro \cite{GoPro} dataset, which contains 2,103 pairs for training and 1,111 pairs for testing. \textbf{v)} Low-light enhancement: we use the LOL \cite{LOL} dataset, which contains 485 pairs for training and 15 pairs for testing. For multi-task image restoration, we train on a mixed dataset containing multiple degradations and test one by one on the dataset containing a single type of degradation. 
\subsubsection{Evaluation Metrics} 
We evaluate performance using reference metrics, including Peak Signal-to-Noise Ratio (PSNR), Structural Similarity Index (SSIM), and Learned Perceptual Image Patch Similarity (LPIPS) \cite{LPIPS}, as well as non-reference metrics such as the Underwater Colour Image Quality Evaluation Metric (UCIQE) \cite{UCIQE}, Underwater Image Quality Measure (UIQM) \cite{UIQM}, 
Multi-scale Image Quality Transformer (MUSIQ) \cite{MUSIQ}, Fog Aware Density Evaluator (FADE) \cite{FADE}, Blind/Referenceless Image Spatial Quality Evaluator (BRISQUE) \cite{BRISQUE}, and Neural Image Assessment (NIMA) \cite{NIMA}. Among these, UCIQE and UIQM are specifically designed for evaluating underwater image restoration tasks, while FADE, BRISQUE, and NIMA are used to assess the performance of dehazing in real-world scenarios. For the metrics PSNR, SSIM, UCIQE, UIQM, MUSIQ, and NIMA, higher scores indicate better performance. In contrast, for the LPIPS, FADE, and BRISQUE metrics, lower scores are preferred. In the tables, the best and second-best scores are highlighted in \textbf{bold} and \underline{underlined}, respectively.

\subsubsection{Baselines} 
In the {All-in-One (``\bf \textcolor{red}{\textbf{N}}+\textcolor{blue}{\textbf{H}}+\textcolor{green}{\textbf{R}}'')} setting, we select four task-specific methods: LPNet \cite{LPNet}, ADFNet \cite{ADFNet}, DehazeFormer \cite{DehazeFormer}, and DRSformer \cite{DRSformer}; five general methods: MPRNet \cite{MPRNet}, Restormer \cite{Restormer}, NAFNet \cite{NAFNet}, FSNet \cite{FSNet}, and MambaIR \cite{MambaIR}, and seven All-in-One methods: DL \cite{DL}, AirNet \cite{AirNet}, IDR \cite{IDR}, NDR \cite{NDR}, PromptIR \cite{PromptIR}, Gridformer \cite{Gridformer} and InstructIR \cite{InstructIR}. 
In the {All-in-One (``N+H+R+B+L'')} setting, we add task-specific methods like HI-Diff \cite{HI-Diff} and Retinexformer \cite{Retinexformer}; general methods such as DGUNet \cite{DGUNet}, and MIRNet-v2 \cite{MIRNet_v2}; as well as All-in-One methods like Transweather \cite{Transweather}, and TAPE \cite{TAPE}. 

In the {One-by-One} setting, we focus on adjustments to task-specific methods. For denoising, we employ DnCNN \cite{DnCNN}, FFDNet \cite{FFDNet}, and ADFNet \cite{ADFNet}. For dehazing, we utilize DehazeNet \cite{DehazeNet}, AODNet \cite{AODNet}, FDGAN \cite{FDGAN}, and DehazeFormer \cite{DehazeFormer}. For deraining, we apply UMR \cite{UMR}, MSPFN \cite{MSPFN}, LPNet \cite{LPNet}, and DRSformer \cite{DRSformer}. For image deblurring, we leverage DeblurGAN \cite{DeblurGAN}, Stripformer \cite{Stripformer}, and HI-Diff \cite{HI-Diff}. Lastly, for low-light image enhancement, we incorporate Retinex-Net \cite{LOL}, URetinex \cite{URetinex}, and Retinexformer \cite{Retinexformer}. In addition to these methods, we also report results from some representative general and All-in-One methods trained in the One-by-One setting.

\subsubsection{Implementation Details} 
For the prompt learning stage, we employ the CLIP model with ViT-B/32 as the backbone. We train Restormer \cite{Restormer} for 100K iterations with a learning rate of $2\times10^{-4}$. The number of embedded tokens $N$ in each learnable prompt is set to 16, and the prompt initialization is set to 100K iterations with a learning rate of $4\times10^{-5}$ and the batch size is set to 32. For the restoration stage, we chose Restormer as the restoration backbone of Perceive-IR and DINO-v2 \cite{DINOv2} base version as the semantic guidance module. We adopt a similar setting to the original Restormer. From level-1 to level-4, the numbers of TB and ETB are [4, 6, 6, 8], attention heads in MDTA and PGCA are both [1, 2, 4, 8], and the channel numbers are [48, 96, 192, 384]. We utilize AdamW optimizer with $ \beta_1 = 0.9$ and $ \beta_2 = 0.999$ to optimize the network. The learning rate is set to $2\times10^{-4}$ with a total batch size of 6 for 400K iterations. The weighting parameters for $\mathcal {L}_{total}$ are: $\lambda_{1}$ = 0.1, $\lambda_{2}$ = 0.05, and $\lambda_{3}$ = 0.1. 
Following \cite{PromptIR}, given the substantial variation in data size across tasks (shown in Tab. \ref{tab:dataset}), we adopted task-specific resampling ratios to balance the training sets. Specifically, a resampling ratio of 3 was applied for denoising, 120 for deraining, 5 for deblurring, and 200 for low-light enhancement, while no resampling was performed for dehazing.
All experiments are conducted on 8 NVIDIA GeForce RTX 3090 GPUs. During training, we utilize cropped patches of size 128 $\times$ 128 as input, and random horizontal and vertical flips are applied to augment the training data.

\begin{table*}[tb]
  \caption{Performance comparison in \textbf{All-in-One (``\textcolor{red}{\bf{N}}+\textcolor{blue}{\bf{H}}+\textcolor{green}{\bf{R}}'')} setting with \textcolor{green}{task-specific}, \textcolor{blue}{general}, and \textcolor{red}{All-in-One} image restoration methods. The reported results are partially based on NDR \cite{NDR} and PromptIR \cite{PromptIR}. * denotes a model that has been retrained.} 
  \label{tab:three_task}
  \centering
    \renewcommand\arraystretch{1.15}
    \resizebox{\linewidth}{!}{\begin{tabular}{c|c|ccc|c|c|c|c} 
  
    \toprule[1pt]
    \toprule[0.5pt]
    \multirow{2}{*}{\bf Type}
    &\multirow{2}{*}{\bf Method} 
    & \multicolumn{3}{c|}{\textbf{Denoising} (CBSD68\cite{BSD68})}
    & \multicolumn{1}{c|}{\bf Dehazing}
    & \multicolumn{1}{c|}{\bf Deraining}
    & \multirow{2}{*}{\bf Average}
    & \multirow{2}{*}{\bf Params (M)}
    \\ \cline{3-7}  

    {}
    &
    & $\sigma = 15$ & $\sigma = 25$ & $\sigma = 50$  
    & SOTS \cite{SOTS}
    & Rain100L \cite{Rain100L}
    &  & \\

    \hline
    \multirow{4}*{\rotatebox{90}{\color{green} \bf Specific}}  

    &(CVPR'19) LPNet \cite{LPNet}
    & 26.47/0.778  & 24.77/0.748  & 21.26/0.552    & 20.84/0.828   & 24.88/0.784  & 23.64/0.738 
    & \phantom{0}2.84\\ 

    &(AAAI'23) ADFNet* \cite{ADFNet}
    &  33.76/0.929   & 30.83/0.871   & 27.75/0.793     &28.13/0.961    &34.24/0.965   &30.94/{0.904} 
    & \phantom{0}7.65\\

    &(TIP'23) DehazeFormer* \cite{DehazeFormer}
    &33.01/0.914   &30.14/0.858   &27.37/0.779     & 29.58/0.970    & 35.37/0.969  &31.09/0.898  
    & 25.44\\ 

    &(CVPR'23) DRSformer* \cite{DRSformer}
    &33.28/0.921   &30.55/0.862   &27.58/0.786     &29.02/0.968    & 35.89/0.970   &{31.26}/0.902 
    & 33.72\\

    \hline
    \hline
    \multirow{5}*{\rotatebox{90}{\color{blue} \bf General}}
    &(CVPR'21) MPRNet \cite{MPRNet}
    & 33.27/0.920  & 30.76/0.871  & 27.29/0.761    & 28.00/0.958   & 33.86/0.958  & 30.63/0.894 & 15.74\\

    &(CVPR'22) Restormer \cite{Restormer}
    & 33.72/{0.930}  & 30.67/0.865  & 27.63/{0.792}    & 27.78/0.958   & 33.78/0.958  & 30.75/0.901 & 26.13\\

    &(ECCV'22) NAFNet \cite{NAFNet}
    & 33.03/0.918  & 30.47/0.865  & 27.12/0.754    & 24.11/0.928   & 33.64/0.956  & 29.67/0.844 & 17.11\\

    & (TPAMI'23) FSNet* \cite{FSNet}
    & 33.81/0.930   &30.84/0.872   & 27.69/0.792     & 29.14/0.968    & 35.61/0.969   & 31.42/0.906  & 13.28 \\
    
    & (ECCV'24) MambaIR* \cite{MambaIR}
    & 33.88/0.931  &  30.95/0.874  & 27.74/0.793    &29.57/0.970   & 35.42/0.969 & 31.51/0.907 & 26.78 \\  

    \hline
    \hline
    \multirow{9}*{\rotatebox{90}{\color{red} \bf All-in-One}}
    &(TPAMI'19) DL \cite{DL}
    & 33.05/0.914  & 30.41/0.861  & 26.90/0.740    & 26.92/0.391   & 32.62/0.931  & 29.98/0.875 
    & \phantom{0}2.09 \\

    &(CVPR'22) AirNet \cite{AirNet}
    & 33.92/{0.932}  & 31.26/{0.888}  & 28.00/0.797    & 27.94/0.962   & 34.90/0.967  
    & 31.20/0.910 & \phantom{0}8.93\\

    &(CVPR'23) IDR* \cite{IDR}
    & {33.89}/0.931  & {31.32}/0.884  & {28.04}/0.798    & 29.87/0.970   & 36.03/0.971  & 31.83/0.911 
    & 15.34 \\

    &(ArXiv'23) ProRes \cite{ProRes}
    & {32.10}/0.907  & {30.18}/0.863  & {27.58}/0.779    & 28.38/0.938   
    & 33.68/0.954  & 30.38/0.888 & 370.63\phantom{0}\\
    
    &(NeurIPS'23) PromptIR \cite{PromptIR}
    & 33.98/\underline{0.933}  & 31.31/{0.888}  & 28.06/{0.799}    
    & \underline{30.58}/\underline{0.974}   & {36.37}/{0.972}  
    & {32.06}/\underline{0.913} & 32.96\\

    &(TIP'24) NDR \cite{NDR}
    & {34.01}/0.932  & {31.36}/0.887  & {28.10}/0.798    & 28.64/0.962   
    & 35.42/0.969  & 31.51/0.910 & 28.40\\

    & (IJCV'24) Gridformer* \cite{Gridformer}
    & 33.93/0.931  & 31.37/0.887  & 28.11/0.801    
    & 30.37/0.970   & 37.15/0.972  
    & 32.19/0.912 & 34.07\\

    & (ECCV'24) InstructIR-3D \cite{InstructIR}
    & \textbf{34.15}/\underline{0.933}  &\underline{31.52}/\bf{0.890}  & \underline{28.30}/\bf{0.804}    
    & {30.22}/{0.959}   & \underline{37.98}/\underline{0.978}  
    & \underline{32.43}/\underline{0.913} & 15.84\\

    & \cellcolor{my_color}\textbf{Perceive-IR (Ours)}  
    & \cellcolor{my_color}\underline{34.13}/\bf 0.934 & \cellcolor{my_color}\textbf{31.53/0.890}  
    & \cellcolor{my_color}\textbf{28.31}/\bf{0.804}  & \cellcolor{my_color}\textbf{30.87/0.975}   
    & \cellcolor{my_color}\textbf{38.29/0.980}  & \cellcolor{my_color}\textbf{32.63/0.917}
    & \cellcolor{my_color}42.02 \\
    
  \bottomrule[1pt]
  \end{tabular}}
\end{table*}

\begin{figure*}[!tbp]  
	\centerline{\includegraphics[page=1,trim = 0mm 0mm 0mm 0mm, clip, width=1\linewidth]{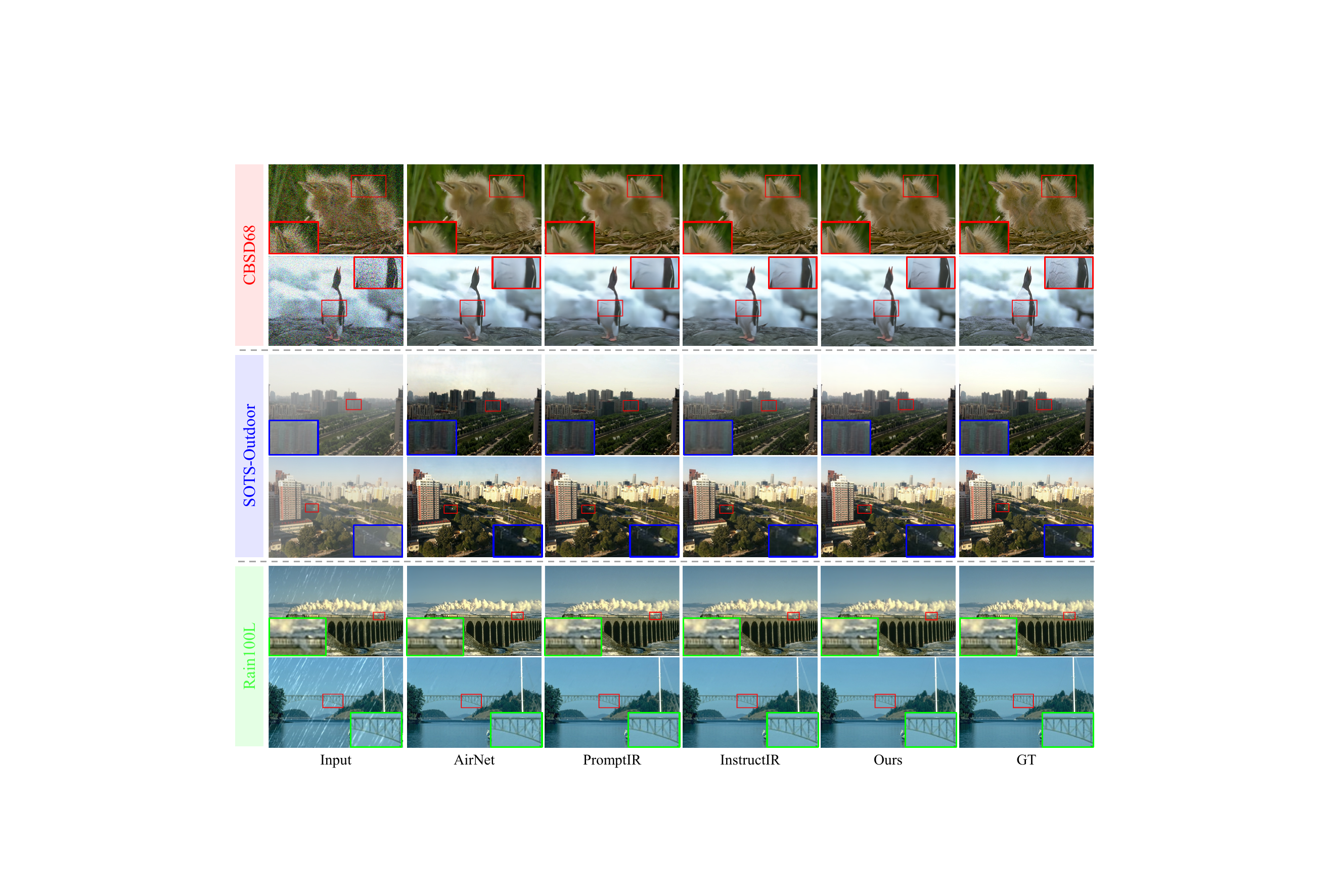}}
	\caption{Visual comparisons of Perceive-IR with state-of-the-art All-in-One methods under ``\textcolor{red}{\bf{N}}+\textcolor{blue}{\bf{H}}+\textcolor{green}{\bf{R}}'' setting. {Zoom-in for best view.}}
	\label{fig:NHR}
\end{figure*}

\begin{table*}[tb]
  \caption{Performance comparison in \textbf{All-in-One (``(\textcolor{red}{\textbf{N}}+\textcolor{blue}{\textbf{H}}+\textcolor{green}{\bf{R}}+\textcolor{purple}{\textbf{B}}+\textcolor{pink}{\textbf{L}})'')} setting with state-of-the-arts \textcolor{green}{task-specific}, \textcolor{blue}{general}, and \textcolor{red}{All-in-One} image restoration methods. Follow \cite{IDR}, denoising results are reported for the noise level $\sigma$ = 25. The reported results are partially based on IDR \cite{IDR}. * denotes a model that has been retrained.}
  \label{tab:five_task}
  \centering
  \renewcommand\arraystretch{1.15}
    \resizebox{\linewidth}{!}{\begin{tabular}{c|c|c|c|c|c|c|c|c} 
  
    \toprule[1pt]
    \toprule[0.5pt]
    \multirow{2}{*}{\bf Type}
    
    & \multirow{2}{*}{\bf Method} 
    & \multicolumn{1}{c|}{\bf Denoising}
    & \multicolumn{1}{c|}{\bf Dehazing}
    & \multicolumn{1}{c|}{\bf Deraining}
    & \multicolumn{1}{c|}{\bf Deblurring}
    & \multicolumn{1}{c|}{\bf Low-Light}
    & \multirow{2}{*}{\bf Average}
    & \multirow{2}{*}{\bf Params (M)}
    \\ \cline{3-7} 

    {} 
    &
    & CBSD68 \cite{BSD68} & SOTS \cite{SOTS} & Rain100L \cite{Rain100L} & GoPro \cite{GoPro} & LOL \cite{LOL}
    &  \\

    \hline
    \multirow{5}*{\rotatebox{90}{\color{green} \bf Specific}}  

    &(AAAI'23) ADFNet* \cite{ADFNet}
    &  31.15/0.882   &24.18/0.928   &32.97/0.943     &25.79/0.781    &21.15/0.823   & 27.05/0.871
    & \phantom{0}7.65\\

    &(TIP'23) DehazeFormer* \cite{DehazeFormer}
    &30.89/0.880   & 25.31/0.937   &  33.68/0.954     &25.93/0.785    &21.31/0.819   & 27.42/0.875 
    & 25.44\\ 

    &(CVPR'23) DRSformer* \cite{DRSformer}
    &30.97/0.881   &24.66/0.931   &33.45/0.953     &25.56/0.780    &21.77/0.821   & 27.28/0.873
    & 33.72\\

    &(NeurIPS'23) HI-Diff* \cite{HI-Diff}  
    &30.61/0.878   &25.09/0.935   &33.26/0.951     & 26.48/0.800    &22.01/0.822   &  27.49/0.877
    & 23.99\\

    &(ICCV'23) Retinexformer* \cite{Retinexformer}  
    & 30.84/0.880  &24.81/0.933   &32.68/0.940     &25.09/0.779    &  22.76/0.834   & 27.24/0.873
    & \phantom{0}1.61\\

    \hline
    \hline
    \multirow{6}*{\rotatebox{90}{\color{blue} \bf General}} 
    
    & (TPAMI'22) MIRNet-v2 \cite{MIRNet_v2}
    & 30.97/0.881  & 24.03/0.927  & 33.89/0.954    & 26.30/0.799   & 21.52/0.815  & 27.34/0.875  
    & \phantom{0}5.86 \\
    
    & (CVPR'22) DGUNet \cite{DGUNet}
    & 31.10/0.883  & 24.78/0.940  &  {36.62}/{0.971}    & 27.25/0.837   & 21.87/0.823  & 28.32/0.891  & 17.33 \\ 

    & (CVPR'22) Restormer \cite{Restormer}
    & {31.49}/0.884  & 24.09/0.927  & 34.81/0.962    & 27.22/0.829   & 20.41/0.806  & 27.60/0.881 & 26.13 \\

    & (ECCV'22) NAFNet \cite{NAFNet}
    & 31.02/0.883  & 25.23/0.939  & 35.56/0.967    & 26.53/0.808   & 20.49/0.809  & 27.76/0.881 & 17.11 \\

    & (TPAMI'23) FSNet* \cite{FSNet}
    & 31.33/0.883  &  25.53/0.943  & 36.07/0.968    & 28.32/0.869   & 22.29/0.829  &  28.71/0.898 & 13.28 \\

    & (ECCV'24) MambaIR* \cite{MambaIR}
    & 31.41/0.884  &  25.81/0.944  & 36.55/0.971    & 28.61/0.875   & 22.49/0.832  & 28.97/0.901  &26.78  \\

    \hline
    \hline
    \multirow{9}*{\rotatebox{90}{\color{red} \bf All-in-One}}
    & (TPAMI'19) DL \cite{DL}
    & 23.09/0.745  & 20.54/0.826  & 21.96/0.762    & 19.86/0.672   & 19.83/0.712  & 21.05/0.743
    & \phantom{0}2.09\\

    & (ECCV'22) TAPE \cite{TAPE}
    & 30.18/0.855  & 22.16/0.861  & 29.67/0.904    & 24.47/0.763   & 18.97/0.621  & 25.09/0.801
    & \phantom{0}1.07\\

    & (CVPR'22) Transweather \cite{Transweather}
    & 29.00/0.841  & 21.32/0.885  & 29.43/0.905    & 25.12/0.757   & 21.21/0.792  & 25.22/0.836
    & 37.93\\

    & (CVPR'22) AirNet \cite{AirNet}
    & 30.91/0.882  & 21.04/0.884  & 32.98/0.951    & 24.35/0.781   & 18.18/0.735  & 25.49/0.846
    & \phantom{0}8.93\\

    & (CVPR'23) IDR \cite{IDR}
    & \bf{31.60}/\bf{0.887}  & 25.24/0.943  & 35.63/0.965    & 27.87/0.846   & 21.34/0.826  & 28.34/0.893 & 15.34 \\

    & (NeurIPS'23) PromptIR* \cite{PromptIR}
    & \underline{31.47}/{0.886}  & {26.54}/{0.949}  & {36.37}/{0.970}    & {28.71}/{0.881}   & {22.68}/{0.832}  & {29.15}/{0.904} & 32.96 \\

    & (IJCV'24) Gridformer* \cite{Gridformer}
    & 31.45/0.885  & 26.79/0.951  & 36.61/0.971    
    & 29.22/0.884   & 22.59/0.831  
    & 29.33/0.904 
    & 34.07\\ 

    & (ECCV'24) InstructIR-5D \cite{InstructIR}
    & {31.40}/\bf 0.887  & \underline{27.10}/\underline{0.956}  & \underline{36.84}/\underline{0.973}    
    &  \underline{29.40}/\bf 0.886   & \bf 23.00/0.836  
    & \underline{29.55}/\underline{0.907} 
    & 15.84 \\

    & \cellcolor{my_color}\textbf{Perceive-IR (Ours)}  
    & \cellcolor{my_color}31.44/\textbf{0.887}  
    & \cellcolor{my_color}\textbf{28.19}/\textbf{0.964}  
    & \cellcolor{my_color}\bf {37.25}/{0.977}    
    & \cellcolor{my_color}\bf{29.46/0.886}  
    & \cellcolor{my_color}\underline{22.88}/\underline{0.833}  
    & \cellcolor{my_color}\textbf{29.84}/\textbf{0.909} 
    & \cellcolor{my_color}42.02 \\
  \bottomrule[1pt]
  \end{tabular}}
\end{table*}

\subsection{All-in-One Restoration Results} 
\label{sect:Multi-task_Results}

Tab. \ref{tab:three_task} presents the overall performance of Perceive-IR and other state-of-the-art methods under the All-in-One ``\textcolor{red}{\bf{N}}+\textcolor{blue}{\bf{H}}+\textcolor{green}{\bf{R}}'' training setting. It can be observed that the All-in-One methods basically perform better than general methods and task-specific methods. Particularly, in terms of overall performance, our Perceive-IR exhibits superior results compared with PromptIR \cite{PromptIR} by over 0.57 dB/0.004 on average in PSNR/SSIM while utilizing the same backbone of Restormer \cite{Restormer}. Compared to the latest methods, Gridformer \cite{Gridformer} and InstructIR \cite{InstructIR}, Perceive-IR also demonstrates a PSNR/SSIM improvement of 0.44 dB/0.005 and 0.20 dB/0.004, respectively.
 
\begin{figure}[!tp]
	\centerline{\includegraphics[page=1,trim = 0mm 0mm 0mm 0mm, clip, width=1\linewidth]{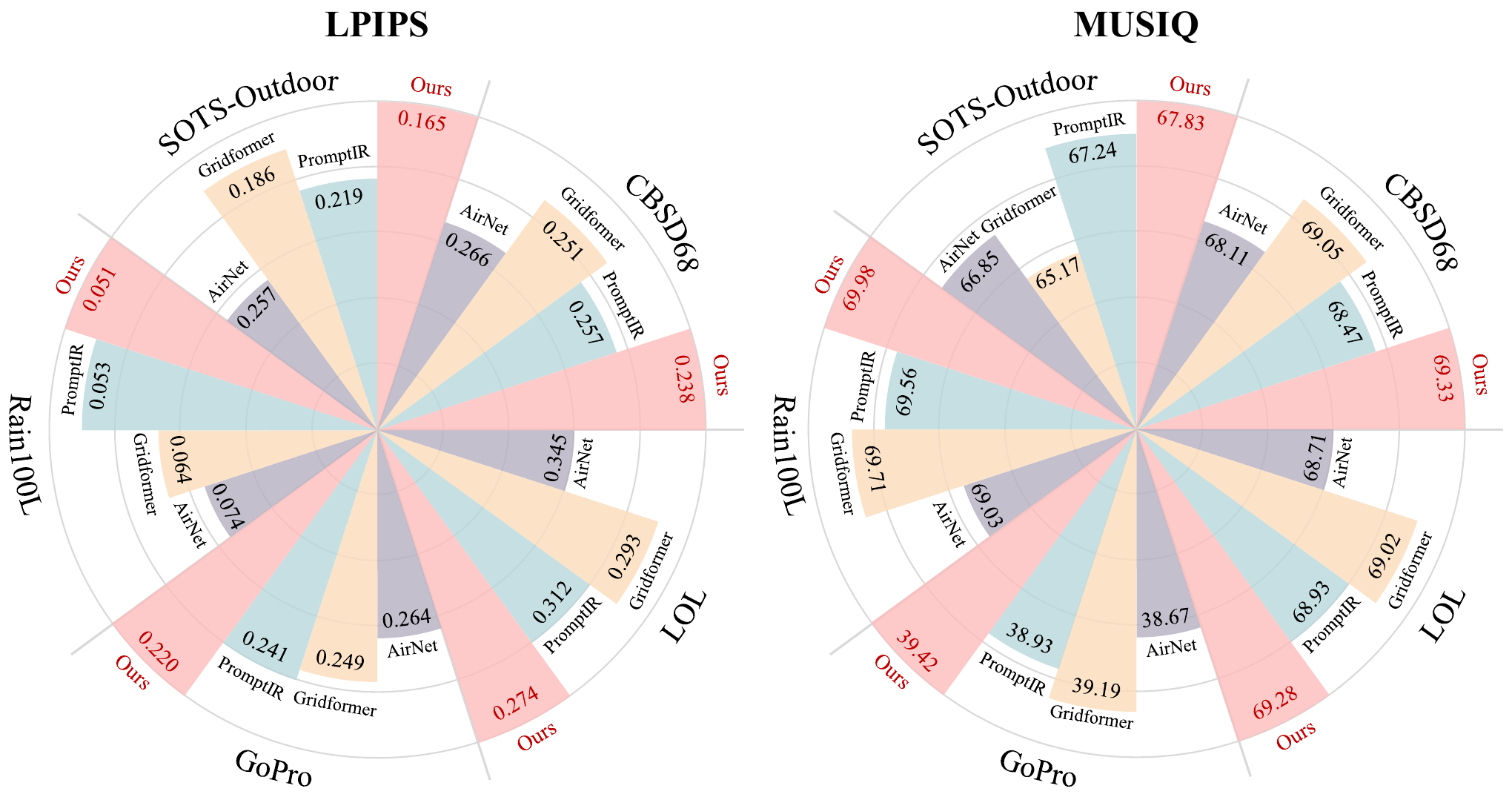}}
	\caption{{A comparison of the LPIPS \cite{LPIPS} and MUSIQ \cite{MUSIQ} perceptual metrics for the All-in-One methods under ``\textbf {\textcolor{red}{\textbf{N}}+\textcolor{blue}{\textbf{H}}+\textcolor{green}{\textbf{R}}+\textcolor{purple}{\textbf{B}}+\textcolor{pink}{\textbf{L}}}'' setting.}}
	\label{fig:LPIPS}
	\vspace{-1em}
\end{figure}

When extending to the more challenging ``\textcolor{red}{\textbf{N}}+\textcolor{blue}{\textbf{H}}+\textcolor{green}{\bf{R}}+\textcolor{purple}{\textbf{B}}+\textcolor{pink}{\textbf{L}}'' setting, as shown in Tab. \ref{tab:five_task}, Perceive-IR continues to demonstrate its superiority. Compared to PromptIR, Gridformer, and InstructIR, it achieves average PSNR/SSIM improvements of 0.69 dB/0.005, 0.51 dB/0.005, and 0.29 dB/0.002, respectively. Interestingly, as the number of degradation types increases, the performance gap between All-in-One methods and general image restoration methods narrows. By contrast, Perceive-IR still maintains good performance.
Additionally, we report the performance of All-in-One methods on perceptual metrics, such as LPIPS \cite{LPIPS} and MUSIQ \cite{MUSIQ}. As illustrated in Fig. \ref{fig:LPIPS}, Perceive-IR achieves the lowest LPIPS scores (indicating higher perceptual similarity) and the highest MUSIQ scores (indicating better overall quality) across all datasets. For example, on the Rain100L dataset, Perceive-IR improves LPIPS by 12.5\% and MUSIQ by 8.3\% compared to the second-best method. These results highlight Perceive-IR's ability to restore images with both high perceptual fidelity and overall quality, validating its effectiveness in diverse restoration scenarios.

\begin{table}[tb] %
  \caption{Denoising performance comparison in the \textbf{One-by-One} setting on \textbf{denoising} task. \textcolor{green}{$\bf\diamond$}, \textcolor{blue}{$\bf\diamond$}, and \textcolor{red}{$\bf\diamond$} denote task-specific, general, and All-in-One image restoration methods, respectively. The results in the table are mainly based on IDR \cite{IDR}.} 
  \label{tab:single_denoising}
  \centering
  \renewcommand\arraystretch{1.05}
  \tabcolsep=0.1cm
    \resizebox{\linewidth}{!}{\begin{tabular}{l|ccc|ccc|ccc} 
  
        \toprule[1pt]
    \toprule[0.5pt]

    \multirow{2}{*}{\bf Method} 
    & \multicolumn{3}{c|}{\textbf{CBSD68} \cite{BSD68}}
    & \multicolumn{3}{c|}{\textbf{Urban100} \cite{Urban100}}  
    & \multicolumn{3}{c}{\textbf{Kodak24} \cite{Kodak24}}\\
    \cline{2-10} 

    {} 
    & 15 & 25 & 50     
    & 15 & 25 & 50 
    & 15 & 25 & 50\\

    \hline

    \textcolor{green}{$\bf\diamond$} DnCNN \cite{DnCNN}
    & 33.90  & 31.24  & 27.95    
    & 32.98  & 30.81  & 27.59    
    & 34.60  & 32.14  & 28.95\\

    \textcolor{green}{$\bf\diamond$} FFDNet \cite{FFDNet}
    & 33.87  & 31.21  & 27.96    
    & 33.83  & 31.40  & 28.05 
    & 34.63  & 32.13  & 28.98 \\ 

    \textcolor{green}{$\bf\diamond$} ADFNet* \cite{ADFNet}
    &  34.21  &  31.60  &  28.19    
    &  34.50  &  32.13  &  28.71   
    &  34.77  & 32.22  &  29.06\\

    \hline
    \hline

    \textcolor{blue}{$\bf\diamond$} MIRNet-v2 \cite{MIRNet_v2}
    & 33.66  & 30.97  & 27.66    
    & 33.30  & 30.75  & 27.22  
    & 34.29  & 31.81  & 28.55 \\

    \textcolor{blue}{$\bf\diamond$} DGUNet \cite{DGUNet}
    & 33.85  & 31.10  & 27.92    
    & 33.67  & 31.27  & 27.94  
    & 34.56  & 32.10  & 28.91 \\ 

    \textcolor{blue}{$\bf\diamond$} Restormer \cite{Restormer}  
    & 34.03  & 31.49  & 28.11    
    & 33.72  & 31.26  & 28.03   
    &  34.78  & 32.37  & 29.08\\  

    \textcolor{blue}{$\bf\diamond$} NAFNet \cite{NAFNet}
    & 33.67  & 31.02  & 27.73    
    & 33.14  & 30.64  & 27.20 
    & 34.27  & 31.80  & 28.62 \\

    \textcolor{blue}{$\bf\diamond$} FSNet* \cite{FSNet}  
    & 34.09   &  31.55   &  28.12    
    & 33.88   &  31.31   &  28.07   
    & 34.75   &  32.38   &  29.10\\

    \hline
    \hline
    \textcolor{red}{$\bf\diamond$} TAPE \cite{TAPE}
    & 32.86  & 30.18  & 26.63
    & 32.19  & 29.65  & 25.87
    & 33.24  & 30.70  & 27.19\\
    
    \textcolor{red}{$\bf\diamond$} AirNet \cite{AirNet}   
    & 34.14  & 31.48  & 28.23   
    & 34.40  & 32.10  & 28.88   
    & \underline{34.81}  & 32.44  & 29.10\\

    \textcolor{red}{$\bf\diamond$} IDR \cite{IDR}
    & 34.11  & 31.60  & 28.14   
    & 33.82  & 31.29  & 28.07 
    & 34.78  & \underline{32.42}  & \underline{29.13} \\

    \textcolor{red}{$\bf\diamond$} PromptIR \cite{PromptIR}
    & \underline{34.34}  & \underline{31.71}  & \underline{28.49}    
    & \underline{34.77}    & \underline{32.49}  & \underline{29.39} 
    & -& -&-\\ 

    \rowcolor{my_color}\textcolor{red}{$\bf\diamond$} \textbf{Perceive-IR}  
    & \textbf{34.38}  & \textbf{31.74}  & \textbf{28.53}    
    & \textbf{34.86}  & \textbf{32.55}  & \textbf{29.42} 
    & \textbf{34.84}  & \textbf{32.50}  & \textbf{29.16}\\
  \bottomrule[1pt]
  \end{tabular}}
  \vspace{-1em}
\end{table}

\begin{table}[htbp] 
  \caption{Performance comparison in the \textbf{One-by-One} setting on {dehazing}, {deraining}, {deblurring}, and {low-light enhancement} tasks. \textcolor{green}{$\bf\diamond$}, \textcolor{blue}{$\bf\diamond$}, and \textcolor{red}{$\bf\diamond$} denote task-specific, general, and All-in-One image restoration methods, respectively. For low-light enhancement, All-in-One methods use 256$\times$256 for training.
  The best overall results are marked with \textcolor{orange}{\textbf{bold}} and the best results from each method category are in \textbf{bold}.}
  \label{tab:single_task2}
  \centering
  \renewcommand\arraystretch{1.15}
  \tabcolsep=0.1cm 
  \resizebox{\linewidth}{!}{\begin{tabular}{lc|lc} 
  
    \toprule[1pt]
    \toprule[0.5pt]

    \multirow{2}{*}{\bf Method} 
    & \cellcolor{Gray}{\textbf{Dehazing}}  
    & \multirow{2}{*}{\bf Method}
    & \cellcolor{Gray}{\textbf{Deraining}}
    \\

    {}
    & \cellcolor{Gray}{SOTS \cite{SOTS}} 
    &
    & \cellcolor{Gray}{Rain100L \cite{Rain100L}}
    \\
    
    \hline
    \textcolor{green}{$\bf\diamond$} DehazeNet \cite{DehazeNet} & 22.46/0.851 
    & \textcolor{green}{$\bf\diamond$} UMR \cite{UMR} & 32.39/0.921 
    \\

    \textcolor{green}{$\bf\diamond$} AODNet \cite{AODNet} & 20.29/0.877 
    & \textcolor{green}{$\bf\diamond$} MSPFN \cite{MSPFN} & 33.50/0.948
    \\

    \textcolor{green}{$\bf\diamond$} FDGAN \cite{FDGAN} & 23.15/0.921
    & \textcolor{green}{$\bf\diamond$} LPNet \cite{LPNet} & 33.61/0.958
    \\

    \textcolor{green}{$\bf\diamond$} DehazeFormer* \cite{DehazeFormer} & \bf \textcolor{orange}{31.78/0.977}  
    & \textcolor{green}{$\bf\diamond$} DRSformer* \cite{DRSformer} & \bf{38.14/0.983}
    \\

    \hline
    \hline
    
    \textcolor{blue}{$\bf\diamond$} Restormer \cite{Restormer} & 30.87/0.969
    & \textcolor{blue}{$\bf\diamond$} Restormer \cite{Restormer} & 36.74/0.978
    \\

    \textcolor{blue}{$\bf\diamond$} NAFNet* \cite{NAFNet}   & 30.98/0.970
    & \textcolor{blue}{$\bf\diamond$} NAFNet* \cite{NAFNet} & 36.63/0.977
    \\

    \textcolor{blue}{$\bf\diamond$} FSNet* \cite{FSNet} & \bf 31.11/0.971
    & \textcolor{blue}{$\bf\diamond$} FSNet* \cite{FSNet} & \bf 37.27/0.980
    \\

    \hline
    \hline
    \textcolor{red}{$\bf\diamond$} AirNet \cite{AirNet} & 23.18/0.900
    & \textcolor{red}{$\bf\diamond$} AirNet \cite{AirNet} & 34.90/0.977
    \\
    
    \textcolor{red}{$\bf\diamond$} PromptIR \cite{PromptIR} & 31.31/0.973
    & \textcolor{red}{$\bf\diamond$} PromptIR \cite{PromptIR} & 37.04/0.979
    \\

    \rowcolor{my_color}\textcolor{red}{$\bf\diamond$} 
    \bf Perceive-IR & \bf 31.65/0.977
    & \textcolor{red}{$\bf\diamond$} 
    \bf Perceive-IR & \bf \textcolor{orange}{38.41/0.984}
    \\
    
  \toprule[1pt]  

    \multirow{2}{*}{\bf Method} 
    & \cellcolor{Gray}{\textbf{Deblurring}}
    & \multirow{2}{*}{\bf Method}
    & \cellcolor{Gray}{\textbf{Low-Light}}
    \\
    
    {}
    & \cellcolor{Gray}{GoPro \cite{GoPro}} 
    &
    & \cellcolor{Gray}{LOL \cite{LOL}}
    \\

    \hline

    \textcolor{green}{$\bf\diamond$} Nah \textit{et al.} \cite{GoPro} & 29.08/0.914 
    & \textcolor{green}{$\bf\diamond$} Retinex-Net \cite{LOL} & 16.77/0.560 
    \\

    \textcolor{green}{$\bf\diamond$} DeblurGAN \cite{DeblurGAN} & 28.70/0.858 
    & \textcolor{green}{$\bf\diamond$} URetinex \cite{URetinex} & 21.33/0.835
    \\

    \textcolor{green}{$\bf\diamond$} Stripformer \cite{Stripformer} & 33.08/0.962 
    & \textcolor{green}{$\bf\diamond$} SMG \cite{SMG} & 23.81/0.809
    \\

    \textcolor{green}{$\bf\diamond$} HI-Diff \cite{HI-Diff} & \bf \textcolor{orange}{33.33/0.964}
    & \textcolor{green}{$\bf\diamond$} Retinexformer \cite{Retinexformer} & \bf \textcolor{orange}{25.16/0.845}
    \\

    \hline
    \hline
    
    \textcolor{blue}{$\bf\diamond$} MPRNet \cite{MPRNet} & 32.66/0.959
    & \textcolor{blue}{$\bf\diamond$} MIRNet \cite{MIRNet} & \bf 24.14/0.835
    \\

    \textcolor{blue}{$\bf\diamond$} Restormer \cite{Restormer} & 32.92/0.961
    & \textcolor{blue}{$\bf\diamond$} Restormer \cite{Restormer} & 22.43/0.823
    \\

    \textcolor{blue}{$\bf\diamond$} FSNet \cite{FSNet} &\bf{33.29/0.963}
    & \textcolor{blue}{$\bf\diamond$} DiffIR \cite{DiffIR} &  23.15/0.828
    \\

    \hline
    \hline
    \textcolor{red}{$\bf\diamond$} AirNet* \cite{AirNet} & 31.64/0.945
    & \textcolor{red}{$\bf\diamond$} AirNet* \cite{AirNet} & 
    {21.52/0.832}
    \\
    
    \textcolor{red}{$\bf\diamond$} PromptIR* \cite{PromptIR} & 32.41/0.956
    & \textcolor{red}{$\bf\diamond$} PromptIR* \cite{PromptIR} & {22.97/0.834}  
    \\

    \textcolor{red}{$\bf\diamond$} Gridformer* \cite{Gridformer} & 
    {32.59/0.957}
    & \textcolor{red}{$\bf\diamond$} Gridformer* \cite{Gridformer} & {23.14/0.829}  
    \\

    \rowcolor{my_color}\textcolor{red}{$\bf\diamond$} 
    \bf Perceive-IR & 32.83/0.960
    & \textcolor{red}{$\bf\diamond$} 
    \bf {Perceive-IR} &  {23.79/0.841}
    \\

    \rowcolor{my_color}\textcolor{red}{$\bf\diamond$} 
    \bf {Perceive-IR (GRL)} & \bf{32.94/0.961}
    & \textcolor{red}{$\bf\diamond$} 
    \bf {Perceive-IR (GRL)} & \bf {23.91/0.843}
    \\
    
  \bottomrule[1pt]

  \end{tabular}}
  \vspace{-1em}
\end{table}

\begin{figure*}[!tbp]  
	\centerline{\includegraphics[page=1,trim = 0mm 0mm 0mm 0mm, clip, width=1\linewidth]{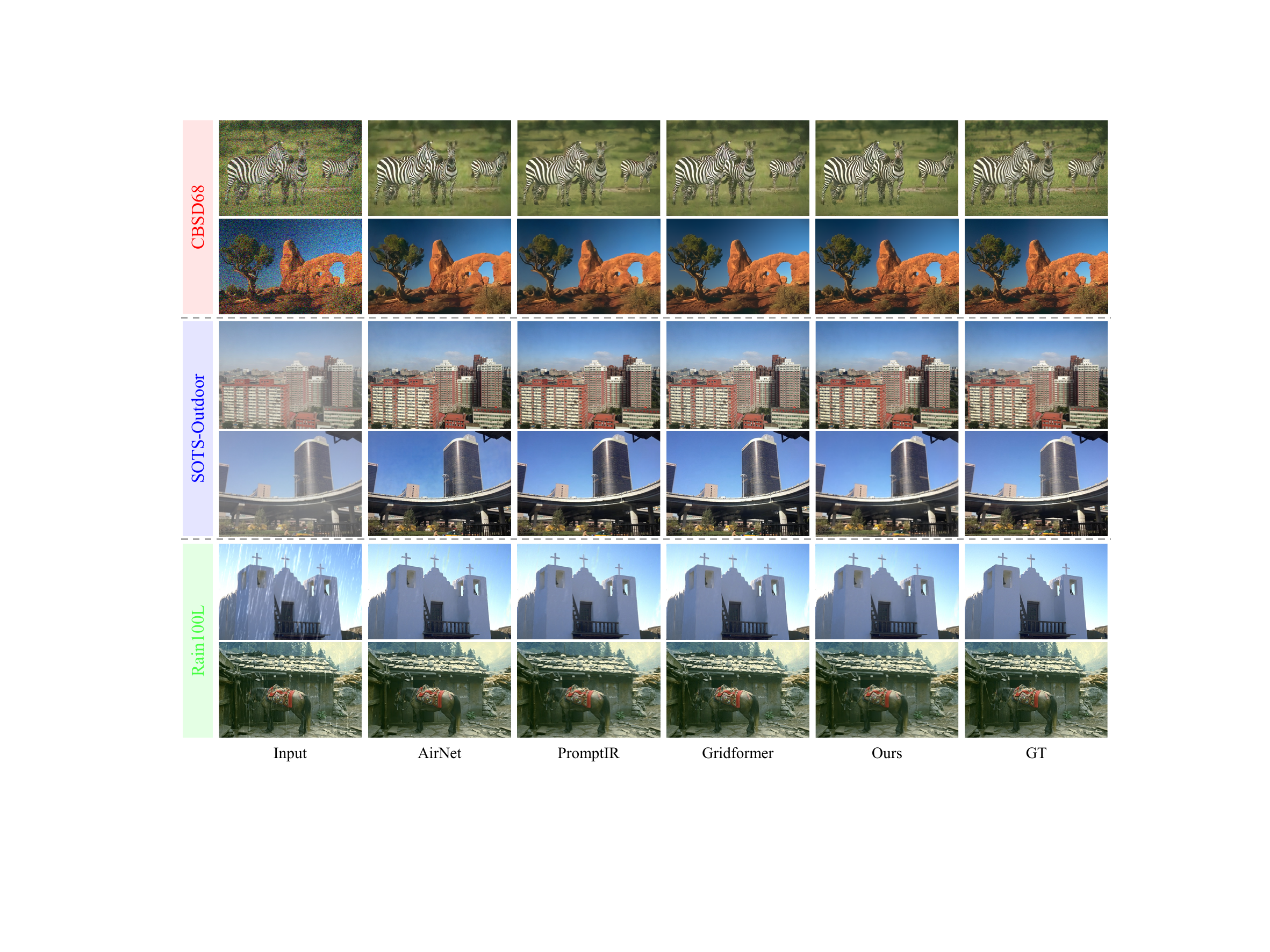}}
	\caption{Visual comparisons of Perceive-IR with state-of-the-art All-in-One methods under One-by-One setting. {Zoom-in for best view.}}
	\label{fig:single}
\end{figure*}

\begin{table*}[tb]
  \caption{Real-world restoration results in \textbf{All-in-One (``\textcolor{red}{\textbf{N}}+\textcolor{blue}{\textbf{H}}+\textcolor{green}{\textbf{R}}'' and ``{\textcolor{red}{\textbf{N}}+\textcolor{blue}{\textbf{H}}+\textcolor{green}{\textbf{R}}+\textcolor{purple}{\textbf{B}}+\textcolor{pink}{\textbf{L}}}'')} settings with state-of-the-art \textcolor{green}{task-specific}, \textcolor{blue}{general}, and \textcolor{red}{All-in-One} image restoration methods. $\dagger$ indicates method has been fine-tuned on the dataset corresponding to the respective real-world scenes.
  Due to the unavailability of weights for InstructIR-3D and InstructIR-5D, we utilize the results of InstructIR-7D for evaluation.}
  \label{tab:generation_real_world}
  \centering
  \tabcolsep=0.1cm 
  \renewcommand\arraystretch{1.25}
\resizebox{\linewidth}{!}{\begin{tabular}{l|c|c|c||c|c|c} 
  
    \toprule[1pt]
    \toprule[0.5pt]
    
    \multirow{3}{*}{\bf Method} 
    & \multicolumn{3}{c||}{\bf \textcolor{red}{\textbf{N}}+\textcolor{blue}{\textbf{H}}+\textcolor{green}{\textbf{R}}}
    & \multicolumn{3}{c}{\bf {\textcolor{red}{\textbf{N}}+\textcolor{blue}{\textbf{H}}+\textcolor{green}{\textbf{R}}+\textcolor{purple}{\textbf{B}}+\textcolor{pink}{\textbf{L}}}}
    \\ \cline{2-7} 

    {} 
    & \cellcolor{my_color1!50}SIDD \cite{SIDD}  
    & \cellcolor{my_color3!50}RTTS \cite{SOTS} 
    & \cellcolor{my_color2}RealRain-1k-L \cite{RealRain-1K}
    & \cellcolor{my_color1!50}SIDD \cite{SIDD}  
    & \cellcolor{my_color3!50}RTTS \cite{SOTS}  
    & \cellcolor{my_color2}RealRain-1k-L \cite{RealRain-1K}
    \\
    \cline{2-7} 
        
    {} 
    & PSNR$\uparrow$/SSIM$\uparrow$   
    & FADE$\downarrow$/BRISQUE$\downarrow$/NIMA$\uparrow$ 
    & PSNR$\uparrow$/SSIM$\uparrow$/LPIPS$\downarrow$
    
    & PSNR$\uparrow$/SSIM$\uparrow$  
    & FADE$\downarrow$/BRISQUE$\downarrow$/NIMA$\uparrow$ 
    & PSNR$\uparrow$/SSIM$\uparrow$/LPIPS$\downarrow$
    \\

    \hline

    Original
    & 23.66/0.439
    & 2.484/36.642/4.483
    & 25.95/0.868/0.407
    & 23.66/0.439
    & 2.484/36.642/4.483
    & 25.95/0.868/0.407
    \\
    \cdashline{1-7}

    \textcolor{green}{$\bf\diamond$} ADFNet \cite{ADFNet}
    & 24.12/0.473   & 1.656/31.254/4.307   &  21.25/0.750/0.414    
    & 24.07/0.470   & 1.738/31.802/4.321   &  20.84/0.746/0.426
\\

    \textcolor{green}{$\bf\diamond$} DehazeFormer \cite{DehazeFormer}
    & 23.89/0.457  & 1.283/24.395/4.504  & 21.92/0.751/0.408
    & 23.76/0.451  & 1.354/28.613/4.552  & 21.68/0.748/0.421
\\ 

    \textcolor{green}{$\bf\diamond$} DRSformer \cite{DRSformer}
    & 23.94/0.461   & 1.464/28.697/4.413  & 22.59/0.773/0.402
    & 23.87/0.453   & 1.488/29.105/4.371  & 22.03/0.755/0.411
\\

    \hline
    \hline
    
    \textcolor{blue}{$\bf\diamond$} FSNet \cite{FSNet}
    & 24.26/0.470   & 1.452/30.681/4.325  & 22.61/0.760/0.404
    & 24.33/0.468   & 1.560/32.964/4.139  & 21.96/0.754/0.408
\\

    \textcolor{blue}{$\bf\diamond$} MambaIR \cite{MambaIR}
    & 24.19/0.465      & 1.326/25.432/4.421 & 22.34/0.754/0.410
    & 23.87/0.457      & 1.494/27.313/4.313 & 21.69/0.751/0.417
\\

    \hline
    \hline

    \textcolor{red}{$\bf\diamond$} AirNet \cite{AirNet}
    & 23.86/0.459      & 1.534/26.845/4.559  & 19.88/0.683/0.401
    & 23.64/0.441      & 1.627/28.345/4.552  & 18.64/0.655/0.409
\\

    \textcolor{red}{$\bf\diamond$} AirNet$\dagger$ \cite{AirNet}
    & 38.34/0.952     
    & 1.198/22.512/4.753  
    & 31.24/0.943/0.183
    & 38.07/0.949   
    & 1.227/23.372/4.673  
    & 30.87/0.938/0.196
\\

    \textcolor{red}{$\bf\diamond$} PromptIR \cite{PromptIR}
    & 24.58/0.482      & 1.298/23.354/4.529 & 22.98/0.767/0.403
    & 24.11/0.469      & 1.356/25.349/4.437 & 22.31/0.759/0.396
\\

    \textcolor{red}{$\bf\diamond$} PromptIR$\dagger$ \cite{PromptIR}
    & 38.73/0.954     
    & \textbf{1.012}/20.607/4.849   
    & 31.69/0.946/0.167
    & 38.66/\textbf{0.954}     
    & 1.185/21.326/4.803  
    & 31.29/0.942/0.178
\\

    \cdashline{1-7}
    \textcolor{red}{$\bf\diamond$} InstructIR-7D \cite{InstructIR}
    & 24.35/0.479      & 1.257/{20.549}/4.573 & 27.21/0.901/0.373
    & 24.35/0.479      & {1.257}/{20.549}/4.573 & 27.21/0.901/0.373
\\
    \cdashline{1-7}
    
    \textcolor{red}{$\bf\diamond$} \cellcolor{my_color}\textbf{Perceive-IR (Ours)}  
    & \cellcolor{my_color}24.88/0.504
    & \cellcolor{my_color}{1.213}/{21.621}/{4.681}
    & \cellcolor{my_color}{27.79}/{0.915}/{0.354}  
    & \cellcolor{my_color}24.65/0.491 
    & \cellcolor{my_color}{1.278}/23.523/{4.624}
    & \cellcolor{my_color}27.43/0.903/{0.370}
\\

    \textcolor{red}{$\bf\diamond$} \cellcolor{my_color}\textbf{Perceive-IR$\dagger$ (Ours)}  
    & \cellcolor{my_color}\bf 39.04/0.954
    & \cellcolor{my_color}1.097/\textbf{18.345}/\textbf{4.901}
    & \cellcolor{my_color}\bf 32.05/0.951/0.144
    & \cellcolor{my_color}\textbf{38.92}/{0.952}
    & \cellcolor{my_color}\bf 1.185/19.021/4.872
    & \cellcolor{my_color}\bf 31.61/0.948/0.160
\\

  \bottomrule[1pt]
  \end{tabular}}
  \vspace{-1em}
\end{table*}

\begin{table}[tbp]
  \caption{Performance comparisons on training-unseen underwater image restoration task. We use the All-in-One methods (pre-trained under \textbf{``{\textcolor{red}{\textbf{N}}+\textcolor{blue}{\textbf{H}}+\textcolor{green}{\textbf{R}}+\textcolor{purple}{\textbf{B}}+\textcolor{pink}{\textbf{L}}}''} training setting) to directly test on the UIEB \cite{UIEB} dataset. The prompt used in InstuctIR \cite{InstructIR} is \emph{``This underwater image is poor, please enhance it.''}.} 
  \label{tab:ablation_generaltion_type_underwater}
  \centering
    \renewcommand\arraystretch{1.15}	
    \resizebox{\linewidth}{!}{\begin{tabular}{l|ccc|cc} 
  
    \toprule[1pt]
  
    \multirow{2}{*}{\bf Method}
    & \multicolumn{3}{c|}{\bf UIEB \cite{UIEB}} 
    & \multicolumn{2}{c}{\bf C60 \cite{UIEB}}
    \\ \cline{2-6} 

    & PSNR$\uparrow$ & SSIM$\uparrow$  & LPIPS$\downarrow$
    & UCIQE$\uparrow$ & UIQM$\uparrow$ \\
    
    \hline
    \hline

    {NAFNet} \cite{NAFNet} 
    & {18.34} & {0.763}  & {0.365}   
    & {0.480} & {1.984}\\

    {Restormer} \cite{Restormer} 
    & {18.69} & {0.782}  & {0.336}   
    & {0.497} & {2.135}\\
    
    {AirNet} \cite{AirNet} 
    & {18.96} & {0.797}  & {0.283}   
    & {0.505} & {2.114}\\
    
    PromptIR \cite{PromptIR} 
    & 20.19 & 0.827  & 0.256   
    & 0.527 & 2.305\\

    Gridformer \cite{Gridformer} 
    & 20.48 &0.846  & 0.243   
    & 0.540 & 2.517 \\

    InstructIR-7D \cite{InstructIR} 
    &21.07  &0.872  &  0.189  
    & 0.552 & 2.489 \\

    \rowcolor{my_color}\bf Perceive-IR 
    & \bf{21.75} & \bf{0.891}  & \bf{0.168}  
    & \bf 0.561 & \bf 2.558\\

  \bottomrule[1pt]
  \end{tabular}}
  \vspace{-1em}
\end{table}

\begin{table}[tbp]
  \caption{Performance comparisons on training-unseen under-display
camera image restoration task. We use the All-in-One methods (pre-trained under \textbf{``{\textcolor{red}{\textbf{N}}+\textcolor{blue}{\textbf{H}}+\textcolor{green}{\textbf{R}}+\textcolor{purple}{\textbf{B}}+\textcolor{pink}{\textbf{L}}}''} training setting) to directly test on the TOLED/POLED \cite{UDC} datasets. The prompt used in InstuctIR \cite{InstructIR} is \emph{``This under-display
camera image is poor, please enhance it.''}.}
  \label{tab:ablation_generaltion_type_UDC}
  \centering
    \renewcommand\arraystretch{1.15}	
    \resizebox{\linewidth}{!}{\begin{tabular}{l|ccc|ccc} 
  
    \toprule[1pt]
  
    \multirow{2}{*}{\bf Method}
    & \multicolumn{3}{c|}{\bf TOLED \cite{UDC}} 
    & \multicolumn{3}{c}{\bf POLED \cite{UDC}}
    \\ \cline{2-7} 

    & PSNR$\uparrow$ & SSIM$\uparrow$  & LPIPS$\downarrow$
    & PSNR$\uparrow$ & SSIM$\uparrow$  & LPIPS$\downarrow$ \\
    
    \hline
    \hline
    {NAFNet} \cite{NAFNet} 
    & {24.90} 
    & {0.828}  
    & {0.360}   
    & {10.68} 
    & {0.475}  
    & 0.713\\

    {Restormer} \cite{Restormer} 
    & {25.74} 
    & {0.801}  
    & {0.338}   
    & {13.94} 
    & {0.428}  
    & 0.681\\
    
    {AirNet} \cite{AirNet} 
    & {25.47} 
    & {0.794}  
    & {0.316}   
    & {12.97} 
    & {0.462}  
    & 0.714\\
    
    PromptIR \cite{PromptIR} 
    & 27.23 & 0.811  & 0.319   
    & 14.54 & \bf 0.493  & 0.728\\

    Gridformer \cite{Gridformer} 
    & 26.17 &0.826  & 0.337   
    & 14.09 & 0.452 & 0.751\\

    InstructIR-7D \cite{InstructIR} 
    & 25.85  &0.843  &  0.305  
    & 13.87 & 0.431  & 0.652\\

    \rowcolor{my_color}\bf Perceive-IR 
    & \bf{29.17} & \bf{0.875}  & \bf{0.294}  
    & \bf 15.89 &  0.477    & \bf 0.531\\

  \bottomrule[1pt]
  \end{tabular}}
  \vspace{-1em}
\end{table}

\begin{table}[tbp] 
  \caption{Performance comparisons of All-in-One methods on unseen noise level of $\sigma$ = 60, 100.}
  \label{tab:ablation_generaltion_severity}
  \centering
    \renewcommand\arraystretch{1.1}	
    {\begin{tabular}{l|cc|cc} 
  
    \toprule[1pt]
  
    \multirow{2}{*}{\bf Method}
    & \multicolumn{2}{c|}{\bf CBSD68 \cite{BSD68}}
    & \multicolumn{2}{c}{\bf Urban100 \cite{Urban100}}
    \\ \cline{2-5} 

    & $\sigma=60$  & $\sigma=100$  & $\sigma=60$  & $\sigma=100$ \\
    
    \hline
    \hline
    AirNet \cite{AirNet}
    & 26.01 & 14.29  
    & 25.11 & 14.23  \\

    PromptIR \cite{PromptIR} 
    & 26.71 &20.23  
    & 27.24 & 20.94  \\

    {Gridformer} \cite{Gridformer} 
    & {26.83}    & {20.14}  
    & {27.16}    & {20.85} \\

    \rowcolor{my_color}\bf Perceive-IR 
    & \bf{27.11} & \bf{20.67}  
    & \bf{27.59} & \bf{21.52}  \\

  \bottomrule[1pt]
  \end{tabular}}
  \vspace{-1em}
\end{table}

\begin{table}[!tb] 
  \caption{Effectiveness of different components on the Rain100L \cite{Rain100L} dataset.}
  \label{tab:ablation_architecture} 
  \centering
  \renewcommand\arraystretch{1.1}
    \resizebox{\linewidth}{!}{\begin{tabular}{c|cccc|cc} 
  
    \toprule[1pt]
  
     Index
    &  \multicolumn{1}{c}{Prompt} 
    &  DINO-v2 
    &  PGM  
    &  CFE  
    & \bf PSNR$\uparrow$
    & \bf SSIM$\uparrow$
    \\

    \hline
    (a)   
    &\textcolor{red}{\ding{55}} &\textcolor{red}{\ding{55}}   
    & \textcolor{red}{\ding{55}}  & \textcolor{red}{\ding{55}}
    & 37.15 & 0.975  \\

    \cdashline{1-7}
    (b)
    &\textcolor{green}{\ding{51}}  & \textcolor{red}{\ding{55}}  
    & \textcolor{red}{\ding{55}} & \textcolor{red}{\ding{55}}
    & 37.37 & 0.979  \\

    (c)
    &\textcolor{red}{\ding{55}} &\textcolor{green}{\ding{51}} & \textcolor{red}{\ding{55}} & \textcolor{red}{\ding{55}}
    & 37.32 & 0.979  \\

    (d)
    &\textcolor{red}{\ding{55}} &\textcolor{red}{\ding{55}} &\textcolor{green}{\ding{51}}  & \textcolor{red}{\ding{55}}
    & 37.28 & 0.978  \\

    (e)
    &\textcolor{red}{\ding{55}} &\textcolor{red}{\ding{55}} &\textcolor{red}{\ding{55}}  &\textcolor{green}{\ding{51}}
    & 37.23
    & 0.977  \\

    (f)
    &\textcolor{green}{\ding{51}} & \textcolor{green}{\ding{51}} & \textcolor{red}{\ding{55}} & \textcolor{red}{\ding{55}}
    & 37.52 & 0.981  \\

    (g)
    &\textcolor{red}{\ding{55}} & \textcolor{red}{\ding{55}} & \textcolor{green}{\ding{51}} & \textcolor{green}{\ding{51}}
    & 37.38 & 0.979  \\

    (h)
    &\textcolor{red}{\ding{55}} & \textcolor{green}{\ding{51}} & \textcolor{green}{\ding{51}} & \textcolor{green}{\ding{51}}
    & 37.45 & 0.979  \\

    (i)
    &\textcolor{green}{\ding{51}} & \textcolor{red}{\ding{55}} & \textcolor{green}{\ding{51}} & \textcolor{green}{\ding{51}}
    & 37.50 & 0.980  \\

    \rowcolor{my_color}(j)
    &\textcolor{green}{\ding{51}} & \textcolor{green}{\ding{51}} & \textcolor{green}{\ding{51}} & \textcolor{green}{\ding{51}}
    & \bf 37.76 & \bf 0.982  \\


  \bottomrule[1pt]
  \end{tabular}}
  \vspace{-1em}
\end{table}

\begin{table}[tbp] 
  \caption{Effectiveness of the different loss functions on the Rain100L \cite{Rain100L} dataset.}
  \label{tab:loss}
  \centering

  \tabcolsep=0.4cm 
  \renewcommand\arraystretch{1.1}	
    {\begin{tabular}{c|ccc|cc} 
  
    \toprule[1pt]
  
    Index
    & $\mathcal{L}_{clip}$ 
    & $\mathcal{L}_{dpl}$
    & $\mathcal{L}_{cl}$
    & \bf PSNR$\uparrow$
    & \bf SSIM$\uparrow$
    \\

    \hline
    (a)
    &\textcolor{red}{\ding{55}}&\textcolor{red}{\ding{55}}& \textcolor{red}{\ding{55}} 
    & 37.30 & 0.978   \\

    \cdashline{1-6}
    (b)
    &\textcolor{green}{\ding{51}}&\textcolor{red}{\ding{55}}&\textcolor{red}{\ding{55}}  
    & 37.51 & 0.980   \\

    (c)
    &\textcolor{red}{\ding{55}}&\textcolor{green}{\ding{51}}&\textcolor{red}{\ding{55}} 
    & 37.42 & 0.979   \\

    (d)
    &\textcolor{red}{\ding{55}}&\textcolor{red}{\ding{55}}&\textcolor{green}{\ding{51}} 
    & 37.37 & 0.979    \\

    (e)
    &\textcolor{red}{\ding{55}}&\textcolor{green}{\ding{51}}&\textcolor{green}{\ding{51}} 
    & 37.46 & 0.979    \\

    (f)
    &\textcolor{green}{\ding{51}}&\textcolor{green}{\ding{51}}&\textcolor{red}{\ding{55}} 
    & 37.61 & 0.981   \\

    (g)
    &\textcolor{green}{\ding{51}}&\textcolor{red}{\ding{55}}&\textcolor{green}{\ding{51}} 
    & 37.55 & 0.981   \\

    (h)\cellcolor{my_color}
    &\cellcolor{my_color}\textcolor{green}{\ding{51}}
    &\cellcolor{my_color}\textcolor{green}{\ding{51}}
    &\cellcolor{my_color}\textcolor{green}{\ding{51}}
    & \cellcolor{my_color}\textbf{37.76} & \cellcolor{my_color}\textbf{0.982}  \\

  \bottomrule[1pt]
  \end{tabular}}
  \vspace{-1em}
\end{table}

\subsection{One-by-One Restoration Results} 
\label{sect:Single_Results} 
In this part, we evaluate Perceive-IR under One-by-One setting. As shown in Tab. \ref{tab:single_denoising}, compared to the state-of-the-art (SOTA) task-specific denoising method ADFNet \cite{ADFNet} and the SOTA general image restoration method FSNet \cite{FSNet}, Perceive-IR surpasses them by 0.17 dB and 0.29 dB in PSNR, respectively, at a noise level of 15 on the CBSD68 and Urban100 datasets.
As shown in Tab. \ref{tab:single_task2}, while Perceive-IR demonstrates highly competitive performance in dehazing and deraining tasks, it does not perform as well as the best task-specific methods (\textit{i.e.}, HI-Diff \cite{HI-Diff} and Retinexformer \cite{Retinexformer}) and the best general methods (\textit{i.e.}, FSNet \cite{FSNet} and MIRNet \cite{MIRNet}) on deblurring and low-light enhancement tasks. 
This limitation may stem from the fact that our method does not explicitly incorporate task-specific priors, which limits its ability to capture critical degradation features and adapt to diverse lighting and illumination conditions. Interestingly, when we replaced the restoration backbone with GRL \cite{GRL} for deblurring and low-light enhancement tasks, we observed performance improvements of 0.11 dB and 0.12 dB, respectively. This adjustment not only demonstrates the adaptability of our framework but also underscores the importance of selecting an appropriate restoration backbone. These results suggest that while our method excels in generalizability, further integration of task-specific priors or backbone enhancements could bridge the performance gap in specialized tasks. A more detailed analysis is provided in Sec. \ref{sec: backbone}.

\subsection{{Generalization}} 
In this section, we evaluate the generalization capabilities of various models, focusing on their zero-shot generalization in real-world scenarios as well as their ability to handle unknown degradation types and severity. Specifically, 
we select noise, haze, and rain as benchmark degradation types and evaluate them on their corresponding real-world datasets: SIDD val dataset \cite{SIDD} (for noise), RTTS \cite{SOTS} (for haze), and RealRain-1k-L test datasets \cite{RealRain-1K} (for rain).
Since the RTTS dataset lacks reference images, we additionally employed several non-reference metrics for quantitative comparison, such as FADE \cite{FADE}, BRISQUE \cite{BRISQUE}, and NIMA \cite{NIMA}.

\subsubsection{Zero-shot generalization in real-world scenes} 
We evaluate the comparative methods trained under the ``\textcolor{red}{\textbf{N}}+\textcolor{blue}{\textbf{H}}+\textcolor{green}{\textbf{R}}'' and ``{\textcolor{red}{\textbf{N}}+\textcolor{blue}{\textbf{H}}+\textcolor{green}{\textbf{R}}+\textcolor{purple}{\textbf{B}}+\textcolor{pink}{\textbf{L}}}'' settings, directly assessing their zero-shot generalization in real-world scenarios. As shown in Tab. \ref{tab:generation_real_world}, for noise removal on the SIDD dataset, Perceive-IR achieves 24.88 dB PSNR and 0.504 SSIM, outperforming InstructIR-7D (24.35/0.479) and PromptIR (24.58/0.482), demonstrating its robustness to unpredictable noise patterns. In hazy scenes from RTTS, Perceive-IR delivers superior perceptual quality with a FADE score of 1.213 (3.5\% lower than InstructIR-7D), maintaining a competitive BRISQUE score of 21.621, and excelling in aesthetic assessment with a NIMA score of 4.681 (compared to 4.573 for InstructIR-7D), effectively addressing the common trade-off between distortion correction and visual naturalness in zero-shot setting. 
In addition, we conducted a series of fine-tuning experiments on several All-in-One approaches. Specifically, for real denoising tasks, we cropped 5,000 paired patches from 320 images in the SIDD training dataset; for real dehazing tasks, we cropped 3,925 paired patches from LMHaze \cite{LMHaze} dataset; and for real deraining tasks, we cropped 3,000 paired patches from 784 images in the RealRain-1k-L training dataset. Each task was fine-tuned for 50k iterations. The results show that the performance of all methods was further improved, with significant gains on SIDD and RealRain-1k-L. Our method consistently maintained nearly the best performance. 

\subsubsection{Generalization to unknown degradation type} 
We utilize the All-in-One methods trained under the ``{\textcolor{red}{\textbf{N}}+\textcolor{blue}{\textbf{H}}+\textcolor{green}{\textbf{R}}+\textcolor{purple}{\textbf{B}}+\textcolor{pink}{\textbf{L}}}'' setting and directly evaluate them on the training-unknown underwater and under-display-camera image restoration tasks. As shown in Tab. \ref{tab:ablation_generaltion_type_underwater} and Tab. \ref{tab:ablation_generaltion_type_UDC}, our Perceive-IR significantly outperforms other general and All-in-One methods on these unknown tasks. These results demonstrate the superior generalization capability of Perceive-IR, particularly in handling unknown degradation scenarios.

\subsubsection{Generalization to unknown degradation severity} 
We utilize the All-in-One methods trained under the noise levels of \( \sigma \in \{15, 25, 50\} \) to test on the unseen noise levels of \( \sigma \in \{60, 100\} \). Tab. \ref{tab:ablation_generaltion_severity} shows that our Perceive-IR outperforms other All-in-One methods on unknown noise level task, achieving PSNR scores of 27.11 (vs. 26.83 for Gridformer) on CBSD68 and 27.59 (vs. 27.24 for PromptIR) on Urban100 at \( \sigma = 60 \), and 20.67 (vs. 20.23 for PromptIR) on CBSD68 and 21.52 (vs. 21.52 for PromptIR) on Urban100 at \( \sigma = 100 \). These results demonstrate its superior generalization capability to handle unknown degradation severity.

\begin{figure*}[!tp] 
	\centerline{\includegraphics[page=1,trim = 0mm 0mm 0mm 0mm, clip, width=1\linewidth]{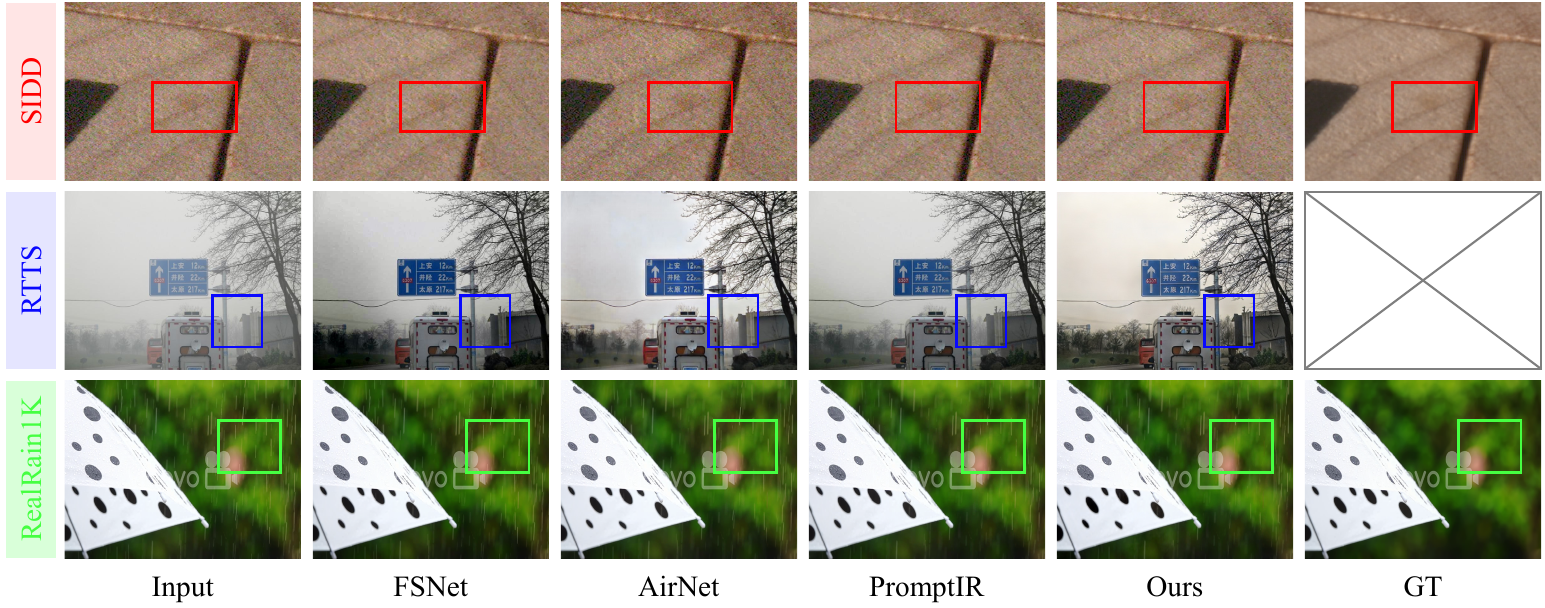}}
	\caption{
    Visual comparisons of Perceive-IR with state-of-the-art All-in-One methods under the ``{\textcolor{red}{\textbf{N}}+\textcolor{blue}{\textbf{H}}+\textcolor{green}{\textbf{R}}+\textcolor{purple}{\textbf{B}}+\textcolor{pink}{\textbf{L}}}'' setting for zero-short generalization in real-world scenes. Zoom-in for best view.}
	\label{fig:real_vis}
	\vspace{-1em}
\end{figure*}

\subsection{Visual Results} 
Figs. \ref{fig:NHR}-\ref{fig:single} present the restoration results obtained by state-of-the-art All-in-One methods for both All-in-One (“\textcolor{red}{\textbf{N}}+\textcolor{blue}{\textbf{H}}+\textcolor{green}{\textbf{R}}”) and One-by-One setting. In noisy scenarios, Perceive-IR produces clear and sharp denoised outputs, preserving fine textures compared to AirNet\cite{AirNet}, PromptIR \cite{PromptIR}, and InstructIR \cite{InstructIR} (\textit{e.g.}, the head and stomach texture parts in rows 1 and 2 of Fig. \ref{fig:NHR}). In challenging hazy scenes, as seen in row 3, AirNet and PromptIR generate results with low color fidelity, while InstructIR produces blurrier results. In row 4, these comparison methods generate results with additional offsets. Conversely, the results generated by our method maintain good color and structural integrity. In deraining task, as shown in the iron-frame region in row 6, AirNet, PromptIR, and InstructIR introduce additional artifacts. In contrast, our approach produced more realistic results. The same trend is illustrated in Fig. \ref{fig:single}. These findings demonstrate the effectiveness of our method. 
{Furthermore, as illustrated in Fig. \ref{fig:real_vis}, although all methods struggle in zero-shot real-world scenarios, our approach still achieves superior visual results. These results underscore the remarkable zero-shot generalization capability of Perceive-IR for real-world image restoration, achieving state-of-the-art performance across a wide range of unknown degradation scenarios.}

\subsection{Ablation Study} 

We conduct several ablation experiments to demonstrate the effectiveness of each component in the proposed Perceive-IR. All ablation experiments are performed on the image deraining task (Rain100L \cite{Rain100L}) by training models for 100K iterations except for special instructions. 

\subsubsection{Effects of Different Components} 
As shown in Tab. \ref{tab:ablation_architecture}, we evaluate the effectiveness of various components through a comparison with the baseline method (index a) that excludes our modules. The baseline adopts Restormer \cite{Restormer} as its backbone. ``Prompt'' denotes the prompt learning stage. 
The settings are as follows: (1) `w/o Prompt' indicates the removal of prompt learning, where the cross-entropy loss ($\mathcal{L}_{ce}$) is omitted; (2) `w/o DINO-v2' refers to replacing the semantic prior $\mathbf{F}_l$ with learnable parameters; (3) `w/o PGM' denotes the direct application of cross-attention using $\mathbf{F}_l$'s features without the processing through $\mathbf{Z}$ and $\mathbf{F}_l$; and (4) `w/o CFE' corresponds to the substitution of $\mathbf{Z}$ with learnable parameters.

Specifically, without prompt learning and DINO-v2 (index g), the average PSNR performance is reduced by 0.38 dB. Similarly, without PGM and CFE (index f), the performance is reduced by 0.24 dB. When compared to other modules, the improvement brought by prompt learning is more significant, as shown in (indexes b-e). This suggests that the proposed CLIP-based prompt learning strategy is crucial to the performance. Overall, our model (index j) achieves a significant average PSNR improvement of 0.61 dB over the baseline, which is attributed to the effectiveness of each proposed component.

\begin{figure}[!tbp]  
	\centerline{\includegraphics[page=1,trim = 0mm 0mm 0mm 0mm, clip, width=1\linewidth]{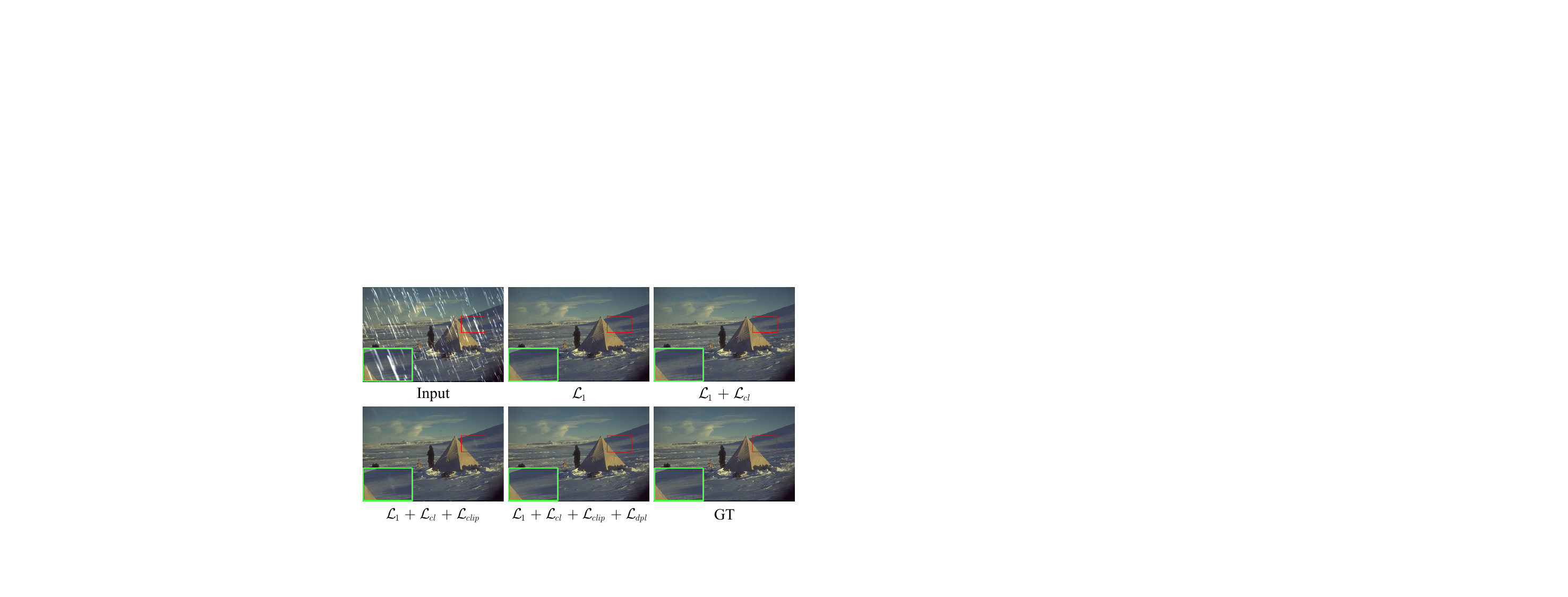}}
	\caption{Visual comparison of the restored images obtained using the individual loss schemes and the proposed scheme.}
	\label{fig:loss_visual}
	\vspace{-1em}
\end{figure}

\begin{figure}[!tbp]  
	\centerline{\includegraphics[page=1,trim = 0mm 0mm 0mm 0mm, clip, width=1\linewidth]{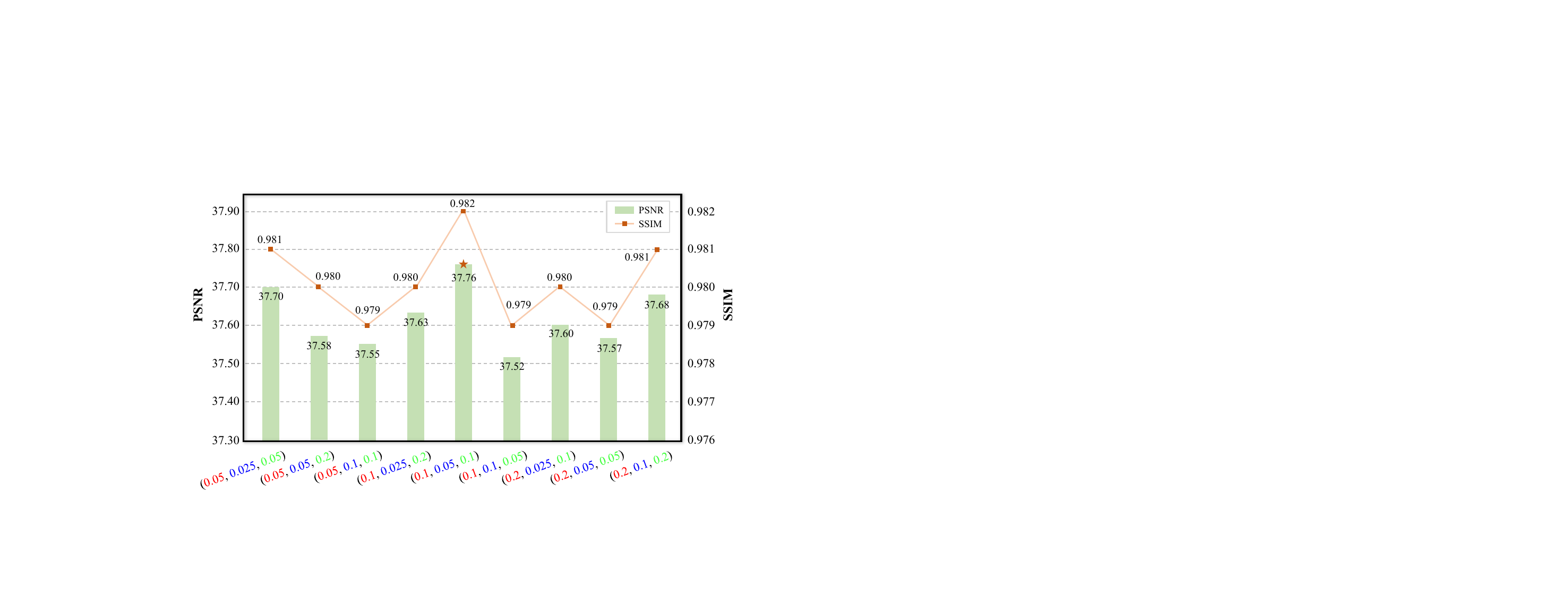}}
	\caption{Comparison of PSNR and SSIM with different weight between different loss function on the Rain100L\cite{Rain100L} dataset. \textcolor{red}{red}, \textcolor{blue}{blue}, and \textcolor{green}{green} are the weights $\lambda_{1}$, $\lambda_{2}$, and $\lambda_{3}$ in Eq. \ref{Eq_totalloss}, respectively.}
	\label{fig:loss_weight}
	\vspace{-1em}
\end{figure}

\subsubsection{Effects of Different Loss Functions} 

As shown in Tab. \ref{tab:loss}, we determine the effectiveness of Perceive-IR under different loss functions. The baseline (index a) uses only pixel-level reconstruction loss $\mathcal{L}_{rec}$. Specifically, compared to the baseline, the approach incorporating $\mathcal{L}_{clip}$, $\mathcal{L}_{dpl}$, and $\mathcal{L}_{cl}$ simultaneously achieves a PSNR improvement of 0.46 dB. Furthermore, by integrating the perceptual capabilities of CLIP, the utilization of $\mathcal{L}_{clip}$ (index b) results in a PSNR improvement of 0.09 dB and 0.14 dB, respectively, over the use of $\mathcal{L}_{dpl}$ (index c) and $\mathcal{L}_{cl}$ (index d) alone. These findings demonstrate that combining multiple loss functions and leveraging CLIP's perceptual properties can significantly elevate the quality of image restoration. Fig. \ref{fig:loss_visual} illustrates the visual results of various schemes using individual loss modules and our proposed scheme. As the proposed loss modules are successively added, the restored image more closely resembles the ground truth.

In addition, we further explored the effect of weights between different loss functions. As shown in Fig. \ref{fig:loss_weight}, peak performance is observed at ($\lambda_{1}$, $\lambda_{2}$, $\lambda_{3}$) = 
(0.1, 0.05, 0.1). Similarly, sub-optimal performance is obtained when $\lambda_{1}$ and $\lambda_{3}$ are larger than $\lambda_{2}$, \textit{e.g.}, (0.05, 0.025, 0.05) and (0.2, 0.1, 0.2). In other cases, the performance degradation is more significant. Thus, in our method, $\lambda_{1}$, $\lambda_{2}$, and $\lambda_{3}$ are set as 0.1, 0.05, and 0.1, respectively.

\subsubsection{Effects of Prompt Number and Initialization Schemes} 

In this study, we examine the impacts of different number of prompts and prompt initialization settings. We consider three scenarios: two, three, and four pairs of prompt-image pairs. Specifically, we explore three configurations for prompt initialization, \textbf{i}) Fixed: The prompts are fixed throughout the entire training process. \textbf{ii}) Random initialization: The prompts are initially randomized and then learned from there. \textbf{iii}) Partial random initialization: Part of the token is initialized randomly while one token is fixed as a specific word. For example, a five-token prompt is initialized as ``$\times$, $\times$, $\times$, $\times$, excellent'', where ``$\times$'' denotes random initialization.  And the entire prompt is updated during prompt learning. 

As shown in Table \ref{tab:ablation_prompt_initialization}, using three prompts in the CLIP model, with the same initialization strategy, outperforms using two or four prompts. The use of two prompt-image pairs may result in inadequate model discrimination, while four pairs could offer too much detail, negatively affecting the portrayal of image quality. This suggests that an appropriate number of prompts is beneficial in providing a more precise guide for the restoration process.

In all prompt configurations, ``Partial random initialization'' consistently outperforms both ``Fixed'' and ``Random initialization''. The use of fixed or random prompts can lead to performance degradation due to the domain gap between the pre-training data of the CLIP model and the specifics of our task. Random prompts slightly enhance performance by easing learning constraints compared with fixed prompts. Therefore, the partial random initialization strategy achieves optimal results, likely because it provides more effective guidance and ensures ample learning space across various prompt scenarios.
We additionally consider the formulation of the specified words. As seen in (indexes g, i) and (indexes l, m), such vague definitions are prone to appear in the data of the pre-trained model, which reduces the performance. When the prompt and image quality do not align, as in (indexes h, i), it disrupts the prompt learning stage, leading to a significant drop in performance. In all, our scheme (index i) achieves the best results and proves its superiority.

\begin{table*}[!tbh]  
  \caption{Performance comparisons on the Rain100L \cite{Rain100L} dataset among different prompt initialization settings for the CLIP model.}
  \label{tab:ablation_prompt_initialization}
  \centering
  \renewcommand\arraystretch{1.18}	
    \resizebox{\linewidth}{!}{\begin{tabular}{l|c|c|cc} %
  
    \toprule[1pt]

    \bf Type
    & Index
    & \bf Prompt Setting
    & \bf PSNR$\uparrow$
    & \bf SSIM$\uparrow$
    \\
    \hline
    \multirow{3}{*}{Two pairs of prompt-image}
    & (a)
    & Fixed (``excellent''/``terrible'')
    & 37.36 & 0.979  
    \\

    \cdashline{2-5}
    & (b)
    & Random initialization
    & 37.49 & 0.980  
    \\

    & (c)
    & Partial random initialization (``excellent''/``terrible'') 
    & 37.56 & 0.981  
    \\ 
    
    \hline
    \hline
    \multirow{6}{*}{Three pairs of prompt-image}
    & (d)
    & Fixed (``excellent''/``mediocre''/``terrible'')
    & 37.45 & 0.979  
    \\

    & (e)
    & Fixed (``good''/``moderate''/``bad'')
    & 37.50 & 0.980  
    \\

    \cdashline{2-5}
    & (f)
    & Random initialization
    & 37.52 & 0.981 
    \\

    & (g)
    & Partial random initialization (``good''/``moderate''/``bad'')
    & 37.68 & 0.982  
    \\

    & (h)
    & Partial random initialization (``terrible''/``moderate''/``excellent'')
    & 37.43 & 0.979   
    \\

    & \cellcolor{my_color}(i)
    & \cellcolor{my_color}Partial random initialization (``excellent''/``mediocre''/``terrible'') (\bf{Ours})
    & \cellcolor{my_color}\bf{37.76} & \cellcolor{my_color}\bf{0.982} \\ 

    \hline
        \hline
    \multirow{4}{*}{Four pairs of prompt-image}
    & (j)
    & Fixed (``excellent''/``slightly better''/``slightly worse''/``terrible'')
    & 37.44 & 0.979  \\

    \cdashline{2-5}
    & (k)
    & Random initialization
    & 37.56 & 0.981 
    \\

    & (l)
    & Partial random initialization (``excellent''/``good''/``bad''/``terrible'')
    & 37.63 & 0.982  
    \\

    & (m)
    & Partial random initialization (``excellent''/``slightly better''/``slightly worse''/``terrible'')
    & 37.67 & 0.982  
    \\

    
  \bottomrule[1pt]
  \end{tabular}}
  \vspace{-1em}
\end{table*}

\begin{table*}[!tbh] 
  \caption{Performance of the proposed Perceive-IR, when conducted on various combinations of degradation types. Number of combinations of tasks from 1 to 5.}
  \label{tab:combination_degradation}
  \centering
  \renewcommand\arraystretch{1.15}
    \resizebox{\linewidth}{!}{\begin{tabular}{c|c|c|c|c||ccc|c|c|c|c} 
  
    \toprule[1pt]
  
    \multicolumn{5}{c||}{\bf Degradation}  
    & \multicolumn{3}{c|}{\textbf{Denoising} (CBSD68 \cite{BSD68})}
    & \multicolumn{1}{c|}{\bf Dehazing}
    & \multicolumn{1}{c|}{\bf Deraining}
    & \multicolumn{1}{c|}{\bf Deblurring}
    & \multicolumn{1}{c}{\bf Low-Light}
    \\
    \cline{1-12} 

    \textcolor{red}{\textbf{N}} 
    & \textcolor{blue}{\textbf{H}} 
    & \textcolor{green}{\textbf{R}} 
    & \textcolor{purple}{\textbf{B}}  
    & \textcolor{pink}{\textbf{L}}
    & $\sigma = 15$ & $\sigma = 25$ & $\sigma = 50$  
    & SOTS \cite{SOTS}
    & Rain100L \cite{Rain100L}
    & GoPro \cite{GoPro}
    & LOL \cite{LOL}
     \\

    \hline
    \textcolor{green}{\ding{51}} 
    &\textcolor{red}{\ding{55}} 
    &\textcolor{red}{\ding{55}} 
    &\textcolor{red}{\ding{55}} 
    &\textcolor{red}{\ding{55}}
    & 34.38/0.939    & 31.74/0.898   & 28.53/0.813    & -  & -  & - & -  \\

    \textcolor{red}{\ding{55}}
    & \textcolor{green}{\ding{51}} 
    & \textcolor{red}{\ding{55}}
    & \textcolor{red}{\ding{55}}
    & \textcolor{red}{\ding{55}}
    &-   &-   &-     & 31.65/0.977   & -  & - & - \\ 

    \textcolor{red}{\ding{55}}
    & \textcolor{red}{\ding{55}}
    & \textcolor{green}{\ding{51}} 
    & \textcolor{red}{\ding{55}}
    & \textcolor{red}{\ding{55}}
    & -  & -  & -    & -   & 38.41/0.984  & - & - \\

    \textcolor{red}{\ding{55}}
    & \textcolor{red}{\ding{55}}
    & \textcolor{red}{\ding{55}}
    & \textcolor{green}{\ding{51}}
    & \textcolor{red}{\ding{55}}
    & -  & -  & -    & -   & -  & 32.83/0.960 & - \\

    \textcolor{red}{\ding{55}} 
    & \textcolor{red}{\ding{55}}
    & \textcolor{red}{\ding{55}}
    & \textcolor{red}{\ding{55}}
    & \textcolor{green}{\ding{51}}
    & -  & -  & -    & -   & -  & - & 23.24/0.838 \\
    
    \hline
    \hline
    \textcolor{green}{\ding{51}} 
    & \textcolor{green}{\ding{51}} 
    & \textcolor{red}{\ding{55}}
    & \textcolor{red}{\ding{55}}
    & \textcolor{red}{\ding{55}}
    & 34.35/0.939 & 31.70/0.897   & 28.46/0.812    & 30.79/0.975   & - & - & -  \\

    \textcolor{green}{\ding{51}} 
    & \textcolor{red}{\ding{55}}
    & \textcolor{green}{\ding{51}} 
    & \textcolor{red}{\ding{55}}
    & \textcolor{red}{\ding{55}}
    & 34.31/0.937  & 31.66/0.895  & 28.41/0.810    & -   & 38.11/0.980  & - & -  \\

    \textcolor{red}{\ding{55}}
    & \textcolor{green}{\ding{51}} 
    & \textcolor{green}{\ding{51}} 
    & \textcolor{red}{\ding{55}}
    & \textcolor{red}{\ding{55}}
    & -  & -  & -    & 30.92/0.976   & 38.23/0.981  & - & -  \\

    \textcolor{red}{\ding{55}}
    & \textcolor{red}{\ding{55}}
    & \textcolor{red}{\ding{55}}
    & \textcolor{green}{\ding{51}} 
    & \textcolor{green}{\ding{51}}
    & -  & -  & -    & -   & -  & 32.33/0.953 & 22.97/0.835  \\

    \hline
    \hline
    \rowcolor{my_color} \textcolor{green}{\ding{51}} 
    & \textcolor{green}{\ding{51}} 
    & \textcolor{green}{\ding{51}} 
    & \textcolor{red}{\ding{55}}
    & \textcolor{red}{\ding{55}}
    & 34.13/0.934  & 31.53/0.890  & 28.31/0.804    & 30.87/0.975   & 38.29/0.980 & - & -   \\

    \textcolor{green}{\ding{51}}     
    &\textcolor{green}{\ding{51}} 
    & \textcolor{red}{\ding{55}}
    &\textcolor{green}{\ding{51}} 
    &\textcolor{red}{\ding{55}}
    & 34.16/0.934  & 31.55/0.890  & 28.32/0.803   & 30.11/0.972   & - & 31.29/0.934 & -   \\

    \textcolor{green}{\ding{51}}     
    & \textcolor{green}{\ding{51}} 
    & \textcolor{red}{\ding{55}}
    & \textcolor{red}{\ding{55}}
    & \textcolor{green}{\ding{51}}
    & 34.20/0.935  & 31.57/0.891  & 28.33/0.804    & 30.36/0.973   & - & - & 22.90/0.834    \\
    
    \hline
    \hline
    \textcolor{green}{\ding{51}}
    & \textcolor{green}{\ding{51}}
    & \textcolor{green}{\ding{51}}
    & \textcolor{green}{\ding{51}}
    & \textcolor{red}{\ding{55}}
    & 34.11/0.932  & 31.50/0.888    & 28.26/0.802   & 28.87/0.967 & 37.83/0.979 & 30.31/0.911& -   \\
    
    \textcolor{green}{\ding{51}}
    & \textcolor{green}{\ding{51}}
    & \textcolor{green}{\ding{51}}
    & \textcolor{red}{\ding{55}}   
    & \textcolor{green}{\ding{51}}
    & 34.13/0.933  & 31.52/0.889  & 28.27/0.802    & 29.62/0.970   & 37.49/0.978 & - & 22.76/0.831   \\

    \textcolor{green}{\ding{51}}     
    & \textcolor{green}{\ding{51}} 
    & \textcolor{red}{\ding{55}}
    & \textcolor{green}{\ding{51}}
    & \textcolor{green}{\ding{51}}
    & 34.10/0.932  & 31.48/0.888  & 28.25/0.801    & 29.23/0.968   & - & 29.89/0.897 & 22.81/0.834   \\

    \hline
    \hline
    \rowcolor{my_color}  
    \textcolor{green}{\ding{51}}
    & \textcolor{green}{\ding{51}}
    & \textcolor{green}{\ding{51}}
    & \textcolor{green}{\ding{51}}
    & \textcolor{green}{\ding{51}}
    & 34.04/0.931  & 31.44/0.887  & 28.19/0.801    & 28.19/0.964   & 37.25/0.977 & 29.46/0.886 & 22.88/0.833   \\
  \bottomrule[1pt]
  \end{tabular}}
  \vspace{-1em}
\end{table*}

\subsubsection{Effects of Different Combinations of Degradation} 
In this study, we assess the performance of Perceive-IR under various combinations of degradation types. As demonstrated in Tab. \ref{tab:combination_degradation}, the performance of the model is optimal on a single task. With the number of tasks increasing, the performance of each task oscillates below the optimal value. In addition, there may be some interesting phenomena under different combinations of degradation. For example, models trained in the ``N+H+L'' setting show superior performance in denoising and dehazing compared to those trained in the ``N+H+B'' setting. Similarly, models trained in the ``N+H+R+L'' setting have better performance on denoising and dehazing than models trained in the ``N+H+R+B'' setting, but worse performance on the deraining task. 
This may be because the distribution of low-light images may contain more noise and haze-like artifacts, which can enhance the model's ability to handle these types of degradation. On the other hand, deraining requires specific features related to rain streaks, which might not be as prevalent in low-light conditions.

\subsubsection{Effects of Restoration Models for Prompt Learning} 
In this part, we examine the impact of restoration methods on prompt learning. As shown in Tab. \ref{tab:diff_IR1}, advanced All-in-One models like PromptIR \cite{PromptIR} underperform AirNet \cite{AirNet}, while general models like Restormer \cite{Restormer} surpass MPRNet \cite{MPRNet}. This can be explained by two factors. First, PromptIR's high-quality outputs exceed the ``mediocre" standard, causing ambiguity in prompt learning. Second, such high-quality restorations may cause the ``mediocre" prompt to converge towards ``excellent", reducing the distinctiveness of the three prompts to essentially two. In contrast, MPRNet's lower-quality results show the opposite effect. Overall, ``medium" quality restorations from appropriate methods are more effective for prompt learning.

\begin{table}[tbp]
  \caption{Performance comparisons on the Rain100L \cite{Rain100L} dataset among different restoration models for the prompt learning stage.}
  \label{tab:diff_IR1}
  \centering
  \renewcommand\arraystretch{1.1}	
    {\begin{tabular}{l|cc} 
  
    \toprule[1pt]
  
    \bf Method
    & \bf PSNR$\uparrow$
    & \bf SSIM$\uparrow$
    \\

    \hline
    
    Perceive-IR (MPRNet \cite{MPRNet})  
    & 37.68 & 0.982   \\

    Perceive-IR (AirNet \cite{AirNet})  
    & 37.64 & 0.981   \\

    Perceive-IR (PromptIR \cite{PromptIR})  
    & 37.57 & 0.981   \\

    \rowcolor{my_color} Perceive-IR (Restormer \cite{Restormer}) (\bf{Ours})
    & \bf{37.76} & \bf{0.982}  \\

  \bottomrule[1pt]
  \end{tabular}}
  \vspace{-1.5em}
\end{table}

\definecolor{lightblue}{rgb}{0.86, 0.95, 1} 
\begin{table}[tbp]
  \caption{Performance comparisons of Perceive-IR with different restoration backbones under the \textbf{``\textcolor{red}{\textbf{N}}+\textcolor{blue}{\textbf{H}}+\textcolor{green}{\textbf{R}}''} and \textbf{``{\textcolor{red}{\textbf{N}}+\textcolor{blue}{\textbf{H}}+\textcolor{green}{\textbf{R}}+\textcolor{purple}{\textbf{B}}+\textcolor{pink}{\textbf{L}}}''} training settings versus state-of-the-art All-in-One image restoration methods. (128, 256) indicates patch size.}
  \label{tab:ablation_backbone}
  \centering
    \renewcommand\arraystretch{1.15}	
    \resizebox{\linewidth}{!}{\begin{tabular}{l|cc|cc|c} 
  
    \toprule[1pt]
  
    \multirow{2}{*}{\bf Method}
    & \multicolumn{2}{c|}{\bf \bf \textcolor{red}{\textbf{N}}+\textcolor{blue}{\textbf{H}}+\textcolor{green}{\textbf{R}} }
    & \multicolumn{2}{c|}{\bf \bf {\textcolor{red}{\textbf{N}}+\textcolor{blue}{\textbf{H}}+\textcolor{green}{\textbf{R}}+\textcolor{purple}{\textbf{B}}+\textcolor{pink}{\textbf{L}}} }
    & \multirow{2}{*}{\bf Params (M)}
    \\ \cline{2-5} 

    & PSNR$\uparrow$  & SSIM$\uparrow$  & PSNR$\uparrow$  & SSIM$\uparrow$ & \\
    
    \hline
    \hline
    AirNet \cite{AirNet}
    & 31.20 & 0.910  
    & 25.49 & 0.846 & 8.93  \\

    PromptIR \cite{PromptIR} 
    & 32.06 & 0.913  
    & 29.15 & 0.904 & 32.96  \\

    InstructIR \cite{InstructIR} 
    & 32.43 & 0.913  
    & 29.55 & 0.907 & 15.84  \\
    
    \hline
    \hline

    \rowcolor{my_color1!50}\textbf{Perceive-IR}$_{\text{NAFNet-128}}$ 
    & {32.52} 
    & {0.915}  
    & {29.71} 
    & {0.908} 
    & {{21.73}}  \\

    \rowcolor{my_color1!50}\textbf{{Perceive-IR}}$_{\text{{NAFNet-256}}}$ 
    & {32.59} 
    & {0.916}  
    & {29.82} 
    & {0.908} 
    &  {{21.73}} \\

    \rowcolor{my_color}\textbf{Perceive-IR}$_{\text{Restormer-128}}$ 
    & {32.63} & {0.917}  
    & {29.84} & {0.909} 
    & {{42.02}}  \\

    \rowcolor{my_color}\textbf{{Perceive-IR}}$_{\text{{Restormer-256}}}$ 
    & {{32.67}} & {{0.917}}  
    & {{29.95}} & {{0.910}} 
    & {42.02}  \\

    \rowcolor{my_color2!50}\textbf{{Perceive-IR}}$_{\text{{X-Restormer-128}}}$ 
    & {{32.66}} & {{0.917}}  
    & {{29.92}} & {{0.909}} 
    & {39.67}  \\

    \rowcolor{my_color2!50}\textbf{{Perceive-IR}}$_{\text{{X-Restormer-256}}}$ 
    & {{32.71}} & {{0.918}}  
    & {{30.04}} & {{0.910}} 
    & {39.67}  \\

    \rowcolor{lightblue!80}\textbf{{Perceive-IR}}$_{\text{{GRL-128}}}$ 
    & {{32.68}} & {{0.917}}  
    & {{29.99}} & {{0.910}} 
    & {35.48}  \\

    \rowcolor{lightblue!80}\textbf{{Perceive-IR}}$_{\text{{GRL-256}}}$ 
    & \bf{{32.72}} & \bf{{0.918}}  
    & \bf{{30.08}} & \bf{{0.911}} 
    & {35.48}  \\

  \bottomrule[1pt]
  \end{tabular}}
\end{table}

\subsection{Discussion and Limitation} 
\label{sec: backbone} 

To validate the adaptability of our method for different restoration backbones, we conducted extensive experiments using NAFNet \cite{NAFNet}, Restormer \cite{Restormer}, X-Restormer \cite{X-Restormer}, and GRL \cite{GRL} as the backbones in the restoration stage. As demonstrated in Tab. \ref{tab:ablation_backbone}, the proposed framework exhibits strong compatibility and superior performance across various architectures. Specifically, Perceive-IR$_{\text{NAFNet-128}}$ achieves 32.52/29.71 dB PSNR under ``N+H+R'' and ``N+H+R+B+L'' settings, surpassing PromptIR \cite{PromptIR} by 0.46/0.56 dB and InstructIR \cite{InstructIR} by 0.09/0.16 dB. Notably, our X-Restormer-128 variant attains state-of-the-art 32.66/29.92 dB PSNR with a 5.6\% parameter reduction compared to the original Restormer backbone (39.67M vs. 42.02M). Consistent performance gains are observed in the GRL-128 variant (\textit{i.e.}, +0.28 dB over the base NAFNet-128, +0.24 dB over the base Restormer-128, and +0.16 dB over the base X-Restormer-128). When trained with a patch size of 256, the performance of all methods improves further, with the relative performance advantages across different backbones remaining intact. These results confirm that our approach serves as a plug-and-play module while enabling scalable integration with advanced models, where the decoupled design allows future works to directly inherit our innovations through backbone upgrades without structural modifications.

Although our Perceive-IR has achieved superior generalization ability and scalability in addressing multiple degradations, its flexibility in blind real-world conditions may be limited. Moreover, while Perceive-IR demonstrates superiority over the state-of-the-art All-in-One and general image restoration methods, it still falls short when compared to the most advanced state-of-the-art task-specific methods, particularly in tasks such as deblurring and low-light enhancement. Lastly, the interrelationships among the various degradation are not sufficiently clear, for example, as shown in Tab. \ref{tab:combination_degradation}, the model performs better on the dehazing task in the ``N+H+R+B'' setting compared to the ``N+H+R+L'' setting, yet the deraining performance is degraded.

\section{Conclusion} 
This paper proposes Perceive-IR, a novel backbone-agnostic framework for All-in-One image restoration that achieves fine-grained quality control through a two-stage process. First, a multi-level quality-driven prompt learning stage trains a quality perceiver to distinguish three-tier quality levels by optimizing prompt-image alignment in the CLIP space. Second, a restoration stage integrates the perceiver with a difficulty-adaptive perceptual loss for quality-aware learning. Additionally, we introduce a Semantic Guidance Module and Compact Feature Extraction to enhance restoration by leveraging semantic priors and degradation-specific features. Extensive experiments demonstrate that Perceive-IR outperforms state-of-the-art methods, generalizes to real-world degradation scenarios, and adapts seamlessly to various backbone networks, highlighting its robustness and flexibility.



%
%

\bibliographystyle{IEEEtran}
\bibliography{IEEEabrv,trans_ref}
\end{document}